\documentclass[nohyperref]{article}

\usepackage{microtype}
\usepackage{graphicx}
\usepackage{subfigure}
\usepackage{enumitem}
\usepackage{booktabs} 

\usepackage{hyperref}

\usepackage[accepted]{icml2023}

\usepackage{amsmath}
\usepackage{amssymb}
\usepackage{mathtools}
\usepackage{amsthm}

\usepackage[capitalize,noabbrev]{cleveref}

\theoremstyle{plain}
\newtheorem{theorem}{Theorem}[section]
\newtheorem{proposition}[theorem]{Proposition}
\newtheorem{lemma}[theorem]{Lemma}
\newtheorem{corollary}[theorem]{Corollary}
\theoremstyle{definition}
\newtheorem{definition}[theorem]{Definition}
\newtheorem{assumption}[theorem]{Assumption}
\crefname{assumption}{assumption}{assumptions}

\theoremstyle{remark}

\newtheorem{example}[theorem]{Example}

\icmltitlerunning{\hfill Asynchronous FL with Bidirectional Quantized Communications and Buffered Aggregation \hfill \thepage}

\usepackage{enumitem}
\usepackage{xspace}
\newcommand{\algname}{{QAFeL}\xspace} 
\newcommand{\norm}[1]{\left\lVert#1\right\rVert}
\newcommand{\expec}[1]{\mathbb{E}\left[#1\right]}

\newcommand{\order}[1]{\mathcal{O}\left(#1\right)}

\newcommand{\calS}{\mathcal{S}}

\DeclareMathOperator{\topk}{top}
\DeclareMathOperator{\randk}{rand}
\DeclareMathOperator{\qsgd}{qsgd}
\DeclareMathOperator{\sign}{sign}

\usepackage[most]{tcolorbox}
\newtcbox{\mybox}[1][]{nobeforeafter,tcbox raise base,colframe=green!50!black,colback=green!10!white,top=0pt,bottom=0pt,left=0pt,right=0pt,before upper=\strut,#1}

\renewcommand{\algorithmicrequire}{\textbf{input:}}
\renewcommand{\algorithmicensure}{\textbf{output:}}

\usepackage{multirow, booktabs}

\begin{document}

\twocolumn[
    \icmltitle{Asynchronous Federated Learning with Bidirectional Quantized Communications and Buffered Aggregation}


    \icmlsetsymbol{equal}{*}

    \begin{icmlauthorlist}
        \icmlauthor{Tomas Ortega}{uci}
        \icmlauthor{Hamid Jafarkhani}{uci}

    \end{icmlauthorlist}

    \icmlaffiliation{uci}{Center for Pervasive Communications \& Computing, University of California, Irvine, USA}

    \icmlcorrespondingauthor{Tomas Ortega}{tomaso@uci.edu}

    \icmlkeywords{Federated Learning, Asynchronous, Quantized Communications, Compressed Communications}

    \vskip 0.3in
]



\printAffiliationsAndNotice{} 

\begin{abstract}
    Asynchronous Federated Learning with Buffered Aggregation (FedBuff) is a state-of-the-art algorithm known for its efficiency and high scalability.
    However, it has a high communication cost, which has not been examined with quantized communications.
    To tackle this problem, we present a new algorithm (\algname), with a quantization scheme that establishes a shared ``hidden'' state between the server and clients to avoid the error propagation caused by direct quantization.
    This approach allows for high precision while significantly reducing the data transmitted during client-server interactions.
    We provide theoretical convergence guarantees for \algname and corroborate our analysis with experiments on a standard benchmark.
\end{abstract}


\section{Introduction}
\label{sec:introduction}

Federated Learning (FL) is a distributed machine learning paradigm that enables training of models on decentralized data, without the need to share raw data~\cite{communication_efficient}.
It has gained significant attention in recent years for its ability to mitigate privacy concerns that come with collecting sensitive information from users in a central location.
Currently, FL is applied to various domains, such as natural language processing, computer vision, and healthcare~\cite{advances_open_problems}.

Many FL algorithms have been widely studied, such as FedAvg~\cite{fedavg_conv_Li}, FedProx~\cite{fedprox}, and FedSGD~\cite{communication_efficient}.
These algorithms operate in a synchronous manner, i.e., all clients send updates to a central server in synchronized rounds.
Since large-scale and dynamic systems are naturally asynchronous \cite{async_hetero}, there is a growing interest in studying asynchronous FL methods, where different clients can update their models and communicate with the server at different times.
While asynchrony introduces additional challenges such as stale gradients and stragglers, it eliminates the need to fit clients into time slots and allows the handling of clients that are slow to respond or have limited communication capabilities~\cite{async_edge}.

Fedbuff~\cite{FedBuff} is an asynchronous FL algorithm that introduces a buffer on the server side to store client updates before performing a global model update.
This is in contrast to previous versions of asynchronous FL, where the server sent a model update every time it received a client update, and the communication cost grew too much with the number of clients.
FedBuff is also compatible with privacy-preserving technologies, has theoretical convergence guarantees under weak assumptions, and is robust to client heterogeneity.
Its superiority over synchronous FL methods in terms of efficiency and fairness has been studied in~\cite{papaya}.
However, its communication cost has not been analyzed in the presence of communication compression.
In this paper, we address this issue with the integration of a bidirectional quantization scheme.
It is of particular interest to investigate the compound error produced by staleness and quantization, which results in a cross-error term that is not present in the separate analysis of both effects.
For more related work, see~\cref{appsec:related-work}.
\vspace{-.1cm}
\paragraph{Contributions.} As our key contributions, we
\begin{itemize}[topsep=0pt, noitemsep]
    \setlength{\itemsep}{0pt}%
    \item Integrate a bidirectional quantization scheme into FedBuff to create a \emph{``hidden'' shared state} between the server and clients and \emph{reduce communication costs}, while avoiding the error propagation caused by direct quantization. We call this algorithm Quantized Asynchronous Federated Learning (\algname{}).
    \item Provide a theoretical analysis of \algname{}'s convergence rate. We show that quantization does not change the complexity order and prove that \emph{FedBuff's rate can be recovered} in the limit of infinite precision quantization.
    \item Show that the \emph{cross-error} term is of smaller order than the individual error from \emph{staleness} and \emph{quantization}.
    \item Validate our findings through an experimental evaluation on a standard benchmark~\cite{LEAF}.
\end{itemize}

\section{System description}
\label{sec:system}
In both \algname and FedBuff, clients train asynchronously and send their local updates to the server when they have finished training.
Simultaneously, the server accumulates local updates in a buffer until it has reached its maximum capacity and then produces a server model update.
In FedBuff, every update is a \emph{full-precision} model.
Current machine learning models can be over tens of millions of parameters large; for example, ResNet-18 has around 11 million trainable parameters~\cite{resnet}.
If floating point numbers are stored in a standard 4 byte format, each client has to upload approximately 44 MB per update.
Since training occurs over thousands of iterations, clients would need to upload information of the order of GB or more.

\begin{figure}[htbp]
    \centering
    \includegraphics[trim=0 .2cm 0 0,clip,width=\linewidth]{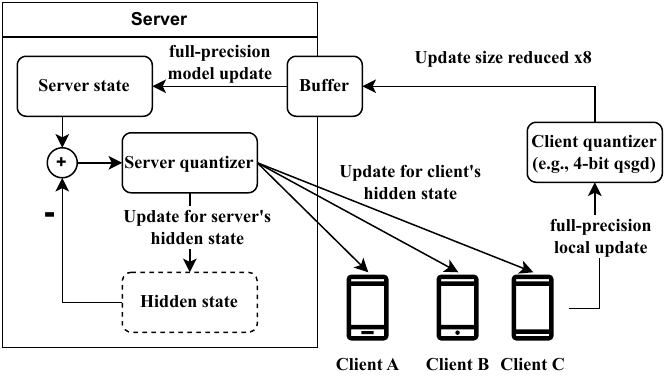}
    \caption{\algname system block diagram, with $\qsgd$ example quantizers. For the definition of a $\qsgd$ quantizer, see \cref{example:quantizers}.}
    \label{fig:system_block_diagram}
\end{figure}

To alleviate this communication cost, \algname \emph{compresses updates} using a quantizer, i.e., a lossy compressor defined as follows~\cite{error_feedback, optimal_compression, new_simpler_ef}.
\begin{definition}[Quantizer] \label{def:quantization}
    A quantizer, denoted by $Q: \mathbb{R}^d \to \mathbb{R}^d$, is a (possibly random) function that satisfies the following condition:
    \begin{equation} \label{eq:quantization}
        \mathbb{E}_Q \left[ \norm{Q(x) - x}^2 \right] \leq (1-\delta) \norm{x}^2,
    \end{equation}
    where $\delta > 0$ is a compression parameter and $\mathbb{E}_Q$ denotes the expectation with respect to the internal randomness of the quantizer $Q$.
    We use the terms quantizer and compression operator indistinctly in this work.
\end{definition}
Using a quantizer allows us to send messages with fewer bits compared to the full model.
\cref{fig:system_block_diagram} sketches the system block diagram in which the server broadcasts the quantized updates.
Both \algname and FedBuff can operate in networks with or without broadcast capabilities.
In this paper, we assume that both operate in the broadcast mode, that is, the server broadcasts the global update once its buffer is full.
For a note regarding the non-broadcast version, see~\cref{appsec:non-broadcast}.

\begin{figure}[htbp]
    \centering
    \includegraphics[trim=0 .1cm 0 0,clip,width=\linewidth]{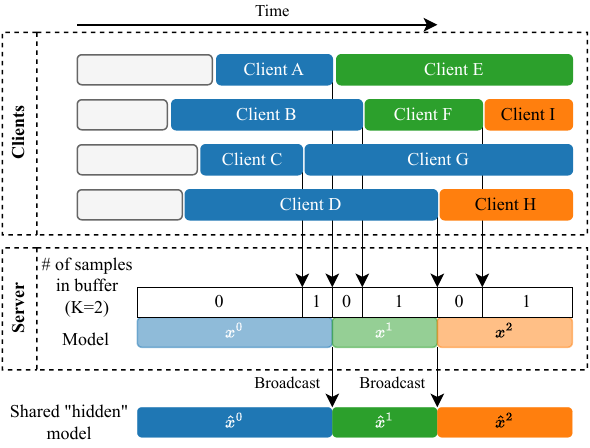}
    \caption{\algname example timeline, with a server with a buffer for $K=2$ samples. Black arrows from clients to the server indicate quantized messages. Black arrows from server to the hidden model indicate a quantized broadcast message. Note that the hidden model is drawn separately from the server and the clients to indicate that it is synchronized.}
    \label{fig:timeline_diagram}
\end{figure}

Let us now describe how \algname works, and illustrate it with an execution timeline example in \cref{fig:timeline_diagram}, similar to the FedBuff analysis done in Figure 4 of \cite{papaya}.
The main difference with respect to FedBuff is that \algname uses a ``hidden'' model, or hidden state, that is shared between the clients and the server.
In practice, this is a model saved in the server's memory and each client, and is used as an approximation to the server model.
Note that the hidden model is different from a direct quantization of the server model, and is used to avoid propagating errors.
To begin training, both the server and the clients start with an initial pre-agreed upon model $x^0$, which is used to initiate the hidden state.
The server then asynchronously samples the clients and requests them to compute a local update.
A requested client will copy the current hidden state, $y_0 \leftarrow\hat x^t$, and perform $P$ local updates using the equation
\begin{equation}
    y_p \leftarrow y_{p-1} - \eta_{\ell} g_p(y_{p-1} ),
\end{equation}
where $\eta_{\ell}$ is the local learning rate and $g_p$ is a noisy gradient.
After the local updates are computed, the client sends the \emph{quantized} difference $Q_c(y_{P-1} - y_0)$ to the server to aggregate.
The server accumulates these updates in a buffer until it has $K$ samples and then performs a global update on the model using the equation
\begin{equation}
    x^{t+1} \leftarrow x^t + \eta_g \frac{\overline{\Delta}^t}{K},
\end{equation}
where $\overline{\Delta}^t$ is the sum of the local updates from the buffer.
The server then updates the hidden state by computing $q^t \leftarrow Q_s(x^{t+1} - \hat x^t)$ and broadcasting it to the clients.
The clients have a process in the background that collects $q^t$.
Then, both the server and the clients perform the hidden state update
\begin{equation}
    \hat x^{t+1} \leftarrow \hat x^t + q^t.
\end{equation}
The full pseudocode for \algname can be found in~\cref{appsec:pseudocode}.
Also, for a note regarding \algname's privacy considerations, see~\cref{appsec:privacy}.

\section{Formulation and convergence analysis}
\label{sec:formulation}
Let us formalize the problem as a minimization of the following sum of stochastic functions:
\begin{equation} \label{eq:minimization_problem}
    \min_{x \in \mathbb{R}^d} f(x) := \frac{1}{N} \sum_{n=1}^N \left( F_n(x) :=  \mathbb{E}_{\zeta_n}[F_n(x;\zeta_n)] \right),
\end{equation}
where $F_n$ is the loss function on Client $n$ and $N$ is the total number of clients.
Each function $F_n$ depends only on data collected locally, i.e., on Client $n$.
Our results can be easily extended to the weighted sum case.

Let us assume that $f$ achieves a minimum value $f^*$.
We make the standard assumptions from the literature \cite{adaptive-fl-optimization, fedavg_conv_Li, stich2018local, Yu-speedup, SCAFFOLD}, which are also used for FedBuff's analysis, \emph{except} the bounded hetereogenity assumption, which is not needed in our proof.
\begin{assumption}[Unbiased stochastic gradient]
    \label{ass:unbiased-stochastic-gradient}
    We assume that for all clients, $\mathbb{E}_{\zeta_n}[g_n(x; \zeta_n))] = \nabla F_n(x)$, $\forall x \in \mathbb{R}^d$.
\end{assumption}
\begin{assumption}[Bounded local variance]
    \label{ass:bounded-local-variance}
    We assume that for all clients, $$\mathbb{E}_{\zeta_n}[\norm{g_n(x; \zeta_n) - \nabla F_n(x)}^2] \leq \sigma^2_{\ell}, \quad \forall x \in \mathbb{R}^d.$$
\end{assumption}
\begin{assumption}[Bounded and L-smooth loss gradient]
    \label{ass:bounded-and-l-smooth-loss-gradient}
    We assume that each function $F_n$ is $L$-smooth, that is,
    $$\norm{\nabla F_n(x) - \nabla F_n(x')} \leq L \norm{x - x'}, \quad \forall x,x'\in \mathbb{R}^d,$$
    and its variance is bounded, i.e., $\norm{\nabla F_n}^2 \leq G$.
\end{assumption}

We also make one additional assumption, which is not standard for synchronous FL, but was introduced in \cite{FedBuff} for the buffered asynchronous setting.
\begin{assumption}[Bounded staleness when $K=1$]
    \label{ass:bounded-staleness}
    For all clients $n\in \{1, \ldots, N\}$ and for each server step $t$, the staleness $\tau_n(t)$, i.e., the difference in model versions between when Client $n$ begins local training and when its updates are applied to the global model, is not greater than a maximum allowed staleness, $\tau_{\max, K}$, when the buffer size $K = 1$.
\end{assumption}
As is the case with FedBuff, it is worth noting that the upper bound on staleness depends on the buffer size, $K$.
As the buffer size increases, the server updates less frequently, which reduces the number of server steps between when a client starts training and when its updates are applied to the global model.
If \cref{ass:bounded-staleness} is met, for any $K>1$, the maximum delay, $\tau_{\max,K}$, is at most $\lceil \tau_{\max,1}/K \rceil$; this is proven in Appendix A of \cite{FedBuff}.

\begin{proposition}[Complexity order.] \label{prop:complexity_order}
    Let us define the convergence rate as $R = \frac{1}{T} \sum_{t=0}^{T-1} \expec{\norm{\nabla f(x^t)}^2}$.
    Under \cref{ass:unbiased-stochastic-gradient,ass:bounded-local-variance,ass:bounded-and-l-smooth-loss-gradient,ass:bounded-staleness}, with local and global learning rates satisfying Condition \eqref{eq:condition_learning_rates} in \cpageref{eq:condition_learning_rates}, $\eta_\ell = \order{1 / (K\sqrt{TP})}$ and $\eta_g = \order{K}$, and defining $F^* := f(x^0) - f^*$,
    \begin{multline}
        R_{FedBuff} = \order{\frac{F^*}{\sqrt{TP}}}  + \order{ \frac{\sigma_{\ell}^2 + G}{K\sqrt{TP}} } \\
        + \order{\frac{ (\tau_{\max, 1}^2 + 1) (\sigma_{\ell}^2 + PG)}{TK^2}}.
    \end{multline}
    Furthermore, with an unbiased client quantizer $Q_c$,
    \begin{equation} \label{eq:R_algname}
        \begin{aligned}
            R_{\algname} & \leq R_{FedBuff} + \order{ \frac{(1-\delta_c)(\sigma_{\ell}^2 + G)}{K\sqrt{TP}} } \\
                         & + \order{\frac{(2-\delta_c)(\sigma_\ell^2 + G)}{\delta_s TK} }                    \\
                         & + \order{\frac{ \tau_{\max, 1}(1-\delta_c) (\sigma_{\ell}^2 + PG)}{TK^2} }.
        \end{aligned}
    \end{equation}
\end{proposition}
Note that the gradient bound $L$ is assimilated into the $\order{\cdot}$ notation, as done in FedBuff's analysis\footnote{The rate that appears in the original AISTATS 2022 paper has a minor error resulted from inaccuracy in Eq. (20) of that paper.}.
\cref{prop:complexity_order}'s proof is in \cref{appsec:convergence-analysis-proof}, where a version for unbiased client quantizers is also outlined.
Moreover, it is also shown how a geometric partial sum in the previous expression is bounded with $1/\delta_s$, but taking the limit without this bound shows that the orders $ \lim_{\delta_c, \delta_s \to 1} R_{\algname} = R_{FedBuff}$.
In other words, \algname recovers FedBuff's convergence rate in the case of infinite precision quantization.

There are three error terms in  \eqref{eq:R_algname}; (i)
the choice of client quantizer with an order $\order{1/\sqrt{T}}$; (ii) the choice of server quantizer with smaller order $\order{1/T}$; and (iii) the cross-error term from client quantization and staleness also with smaller order $\order{1/T}$.
The effects of the server quantizer and the cross-error term dissipate in time faster than the effect of the client quantizer.
Therefore, the choice of the client quantizer will affect the error order more than the choice of the server quantizer and also more than the staleness and quantization cross-error.

\begin{figure*}
    \centering
    \hfill
    \includegraphics[width=.4\textwidth]{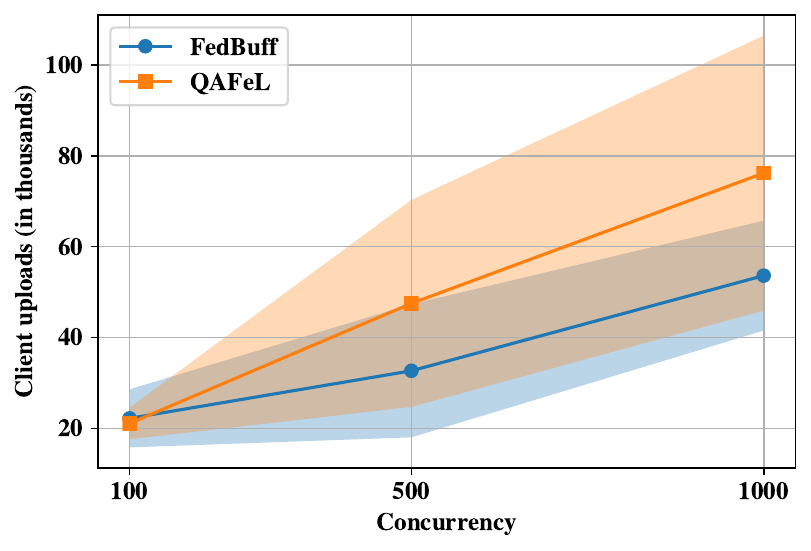}
    \hfill
    \includegraphics[width=.4\textwidth]{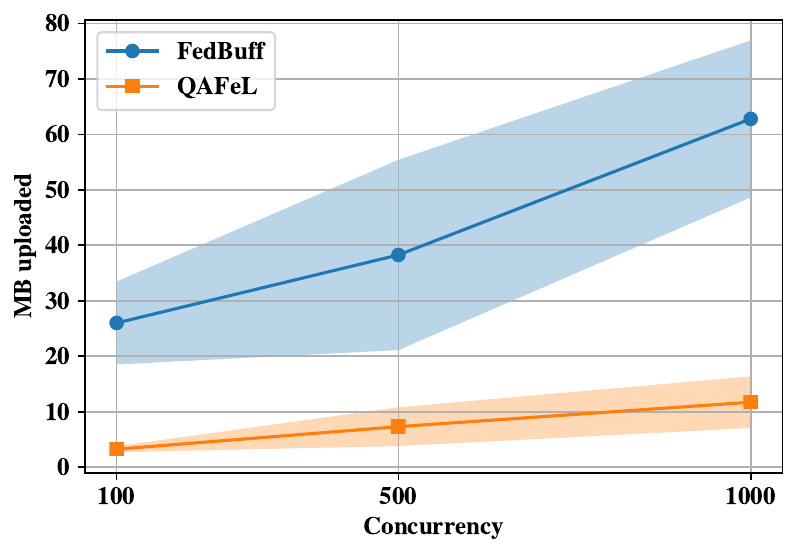}
    \hfill
    \caption{\algname and FedBuff's communication metrics to reach a validation accuracy (90\%) for different concurrency values (clients training in parallel). \algname is using 4-bit $\qsgd$ quantization at both server and client, therefore the MB broadcasted are simply the MB uploaded divided by the buffer size, which is 10.
    }
    \label{fig:Fig3FedBuff}
\end{figure*}
\section{Simulation results}
\label{sec:experiments}

We conduct a series of simulations to evaluate \algname's performance and confirmed our theoretical derivations.
Our simulations show how \algname's communication load is $8$ times smaller than that of FedBuff while maintaining the same convergence speed.

\paragraph{Setup.}
A detailed description of the experimental setup is provided in \cref{appsec:experimental-details}.
The parameters are adopted from~\cite{FedBuff} unless they are not defined in that paper.
The main traits are (i) We use the same hyperparameters as FedBuff, (ii) we model clients arriving at a constant rate, and (iii) to simulate the time delay between a client's download and upload operation, we sample from a half-normal distribution.
This distribution is selected as it provides the most accurate representation of the delay distribution observed in Meta's production FL system, see Appendix C of \cite{FedBuff}.

In addition to the standard metric for comparing synchronous and asynchronous FL methods (aka the number of client trips), we also present the number of bytes sent per message to illustrate \algname's benefits.

\paragraph{Dataset and model.}
We utilize the CelebA dataset and model inherited from \cite{FedBuff}, which follows the configuration of the standard LEAF benchmark \cite{LEAF}.
The CelebA dataset~\cite{celeba} is a large-scale image classification dataset featuring celebrity faces with a total of 202,599 images and 10,177 annotated identities.
The images in CelebA exhibit diverse variations in pose, expression, and appearance.
Our task is to detect whether a celebrity is smiling or not.

Our model's high-level architecture consists of a four-layer convolutional neural network (CNN) binary classifier, with a stride of 1, a padding of 2, and a dropout rate of 0.1.

We use 4-bit $\qsgd$ quantization for both client and server -- see \cref{example:quantizers} for a definition of $\qsgd$ quantization and other examples.
The choice of quantizer follows the results in \cref{fig:threefigures}, where we can see \algname's performance for different combinations of $\qsgd$ quantizers.
The results of \cref{fig:threefigures} are clearly consistent with the analysis presented in \cref{sec:formulation}, as the effect of the server quantizer is less pronounced than that of the client quantizer.

\paragraph{Results.}
\label{par:results}
We perform each experiment three times and report the mean and standard deviation of the client uploads.
\cref{fig:Fig3FedBuff} illustrates how \algname's communication costs are lower than those of FedBuff: we decrease the MB uploaded by  $5.2-8$ times and similarly decrease the MB broadcasted.
The number of client updates is only $1-1.5$ times higher, but using 4-bit $\qsgd$ quantization in both client and server accounts for approximately a $8$ times reduction in message size.
Note that the decrease in the total MB required for upload and broadcast include the extra client updates.
Additional results can be found in~\cref{appsec:additional-results}.

\section{Conclusion}
\label{sec:conclusion}
We study the impact of bidirectional quantization on the convergence rate of buffered asynchronous FL.
Using a hidden state scheme, we avoid error propagation.
We show that the cross-error term is of smaller order compared to the individual terms corresponding to staleness and quantization.
Empirical evaluation corroborates our analysis and shows how client quantization affects \algname's performance more than server quantization.
The theoretical analysis presented in this work can serve as a foundation for the development of quantization schemes that ensure a specific convergence rate, for both biased and unbiased server quantizers.
This approach can be used to design FL systems with bandwidth constraints, which is a common scenario.
\newpage
\bibliography{mybib}
\bibliographystyle{icml2023}

\newpage
\onecolumn
\part*{Appendix}
\appendix
\section{Related work}
\label{appsec:related-work}
Previous work has extended the theoretical understanding of FL to the asynchronous regime \cite{asyncfl, hogwild, improved_async, critical_params, perturbedIterate}.
Most of this work has studied the convergence behavior of asynchronous FL for the setting of homogeneously distributed data, with i.i.d client objective functions.
This is well-known to be unrealistic in many instances of practical FL~\cite{advances_open_problems}.
FedBuff has been proposed as a practical alternative, with buffered aggregation to handle high concurrency settings, as well as being compatible with protection against inference attacks~\cite{FedBuff}.
Sharper convergence rates compared to those of FedBuff are found in \cite{sharper_async}.
They depend only on the average staleness, not on the maximum.
Despite the better rates, this work does not accommodate buffered aggregation, which is highly desirable for scalability.

There is also extensive literature that analyzes synchronous FL with quantized communications; see, for example,~\cite{communication_efficient}, where quantization from client to server was addressed.
Further studies argued the benefits of standard bidirectional quantization, which further reduces the communication cost of FL, and is of importance, particularly in wireless mediums~\cite{quantized_global_model}.
There has also been previous work in distributed and FL with error feedback (or error control), which we use to create a ``hidden'' state in this work~\cite{new_simpler_ef, EF21-P-and-friends, gossiped_OMKL, choco_sgd}.
Unlike this work, the existing literature on quantized FL assumes synchronous FL.

\section{Additional remarks}
\label{appsec:additional-remarks}
\subsection{Non-broadcast variant.} \label{appsec:non-broadcast}
As previously mentioned, \algname can be easily modified to accommodate networks without broadcast capabilities.
In this case, the server must keep the hidden state updates in storage for a maximum of $C_{\max}$ updates, where $C_{\max}$ is the storage size of the model divided by the expected size of a compressed hidden state update.
When the server samples an available client, the server receives the client's hidden state staleness and then transmits the necessary updates so that the hidden state is up to date.
If the staleness is larger than $C_{\max}$, the server simply transmits the hidden state to the client.
In both scenarios, the communication cost of \algname is less than or equal to that of FedBuff.

\subsection{Privacy considerations.} \label{appsec:privacy}
Past research has shown that sensitive information can be recovered from the gradients of participating clients \cite{feature_leakage,inverting_gradients}; this is known as an interference attack.
To protect against this privacy threat, FedBuff uses techniques known as secure aggregation \cite{Cryptonite,SecAgg} and differential privacy \cite{practicalAndPrivate,learning_dp}.
Since our algorithm \algname extends FedBuff, it is also compatible with these techniques and can benefit from the same level of privacy protection.
It is worth noting that our algorithm only requires the server to send perfect updates to the clients, which does not interfere with the privacy scheme of FedBuff.

\subsection{Quantizer examples.}
\begin{example}[Three standard quantizers] \label{example:quantizers}
    Given a vector $x \in \mathbb{R}^d$, let us define
    \begin{itemize}
        \item $\topk_k(x)$ sends the largest $k$ out of the $d$ coordinates of $x$.
        \item $\randk_k(x)$ sends $k$ out of the $d$ coordinates of $x$, chosen at random.
        \item $\qsgd_s(x)$, given a positive integer $s$ that sets the number of quantization levels, sends bits that represent $\norm{x}$, $\sign (x)$, and $\xi(x, s)$,
              \begin{equation*}
                  \xi_i(x,s) =
                  \begin{cases}
                      \lceil \frac{x_i \cdot s}{\norm x} \rceil   & \text{ with probability } \frac{x_i \cdot s}{\norm x} - \lfloor \frac{x_i \cdot s}{\norm x} \rfloor, \\
                      \lfloor \frac{x_i \cdot s}{\norm x} \rfloor & \text{ otherwise},
                  \end{cases}
              \end{equation*}
              and the receiver can reconstruct $\tfrac{\norm{x}}{s} \cdot \sign (x) \cdot \xi(x, s)$.
    \end{itemize}
    For $\topk_k$ and $\randk_k$, their compression parameter $\delta$ is $k/d$, as proven in Lemma A.1 of~\cite{sparsifiedSGD}.
    For $\qsgd$ with $s$ levels of quantization, $\delta = 1 - \min(\tfrac{2d}{s^2}, \tfrac{\sqrt{2d}}{s})$, see Lemma 3.1 in \cite{qsgd}.
    Note also that $\topk_k$ is the only biased quantizer out of the three.
    Nevertheless, there exist methods to transform $\topk_k$ and general biased quantizers into unbiased ones, at the price of extra data transmission, see~\cite{unbiased_horvath}.

    One can easily see how the first two quantizers, $\topk_k$ and $\randk_k$, save data by only sending some components of the vector.
    For $\qsgd$, it is important to note that $\xi_i(x,s)$ are integers that go from 0 to $s$.
    Therefore, if we only use $n$ bits to represent these integers, we can send only $n$ bits per coordinate instead of the full precision floating point number, which is usually 32 bits.
    This is called an $n$-bit $\qsgd$ quantizer and the number of bits per coordinate, $n$, automatically determines the quantization level $s$.
\end{example}

\section{Full pseudocode for \algname}
\label{appsec:pseudocode}
To alleviate the communication cost of asynchronous FL with buffered aggregation, we propose \algname, which consists of three parts: \algname-server, \algname-client, and \algname-client-background.
\algname-server is stated in \cref{alg:server}, which runs in the server and calls \algname-client (\cref{alg:client}) when it needs an update from a client.
In the background, clients are always running \algname-client-background, described in \cref{alg:client-background}.
The highlighted texts mark the lines that differ from FedBuff.
Note that \cref{alg:client-background} is completely new.

\begin{algorithm}[p]
    \caption{\algname-server} \label{alg:server}
    \begin{algorithmic}[1]
        \REQUIRE {server learning rate $\eta_g$, client learning rate $\eta_{\ell}$, client SGD steps $P$, buffer size $K$, initial model $x^0$}
        \STATE \mybox{$\hat x^0 \leftarrow x^0$\hspace{10.68cm} \{initialize shared hidden state\} }
        \REPEAT
        \STATE{$c \leftarrow$ sample available clients \hfill\COMMENT{async}}
        \STATE{run {\algname-client}$(\eta_{\ell}, P)$ on $c$ \hfill\COMMENT{async}}
        \IF{receive client update}
        \STATE{$\Delta_n \leftarrow$ received \mybox{quantized} update from client $i$}
        \STATE{$\overline{\Delta}^t \leftarrow \overline{\Delta}^t + \Delta_n$}
        \STATE{$k \leftarrow k + 1$} \hfill\COMMENT{number of clients in buffer}
        \ENDIF
        \IF{$k == K$}
        \STATE {$\overline{\Delta}^t \leftarrow \frac{\overline{\Delta}^t}{K}$}
        \STATE {$x^{t+1} \leftarrow x^t + \eta_g \overline{\Delta}^t$}
        \STATE {\mybox{Broadcast $q^t \leftarrow Q_s(x^{t+1} - \hat x^t)$}}
        \STATE \mybox{$\hat x^{t+1} \leftarrow \hat x^t + q^t$ \hspace{9.2cm} \{update shared hidden state\} }
        \STATE  {$\overline{\Delta}^t \leftarrow 0, k \leftarrow 0, t \leftarrow t + 1$ \hfill\COMMENT{reset buffer}}
        \ENDIF
        \UNTIL{convergence}
        \ENSURE {FL-trained global model}
    \end{algorithmic}
\end{algorithm}

\begin{algorithm}[p]
    \caption{\texttt{\algname{}-client}} \label{alg:client}
    \begin{algorithmic}[1]
        \REQUIRE client learning rate $\eta_{\ell}$, number of client SGD steps $P$
        \STATE{$y_0 \leftarrow$ \mybox{$\hat x^t$ \hspace{8.95cm} \{availability of $\hat x^t$ ensured by algorithm~\ref{alg:client-background}\} }}
        \FOR{$p=1:P$}
        \STATE{$y_p \leftarrow y_{p-1} - \eta_{\ell} g_p(y_{p-1} )  $}
        \ENDFOR
        \STATE{$\Delta \leftarrow$ \mybox{$ Q_c(y_0 - y_p)$ \hspace{9.38cm} \{Using an ubiased quantizer\}}}
        \STATE{send $\Delta$ to server}
        \ENSURE client update $\Delta$
    \end{algorithmic}
\end{algorithm}

\begin{algorithm}[p]
    \caption{\mybox{\texttt{\algname{}-client-background}}} \label{alg:client-background}
    \begin{algorithmic}[1]
        \REQUIRE {initial model $x^0$}
        \STATE {$\hat x^0 \leftarrow x^0$}
        \REPEAT
        \STATE {wait for quantized update $q^t$}
        \STATE {$\hat x^{t+1} \leftarrow \hat x^t + q^t$}
        \UNTIL{shutdown}
        \ENSURE {updated FL global model}
    \end{algorithmic}
\end{algorithm}

\section{Experimental details}
\label{appsec:experimental-details}
We have utilized the scripts provided by the LEAF benchmark for non-iid client partitions, with a fixed random seed of 1549775860, to divide the users into 80\% training, 10\% validation, and 10\% test sets, respectively.
This gives us 7474, 1869, and 1869 train, validation, and test users, respectively.
Each user has between 1 and 32 samples.
As is standard practice for image datasets, we preprocess the train, validation, and test images.
Specifically, we performed a resize and center crop operation on each image to obtain a resolution of $32 \times 32$ pixels, followed by normalizing each channel of the image to have a mean of $0.5$ and a standard deviation of $0.5$.

We employ a convolutional neural network (CNN) classifier as our model, which is modified from the version used in the LEAF benchmark by replacing batch normalization layers with group normalization layers~\cite{groupnorm, batchnorm_with_groupnorm}.
This modification was carried out to replicate the approach utilized in the experiments performed by FedBuff.
The high-level architecture of our model consists of a four-layer CNN binary classifier, which utilizes a stride of 1, a padding of 2, and a dropout rate of 0.1.

All experiments are run 3 times, and the average is reported, along with the standard deviation.
The hyperparameters we use are: client learning rate $\eta_\ell = 4.7 \cdot 10^{-6}$, server learning rate $\eta_g = 1000$, server momentum $\beta = 0.3$, and buffer size $K = 10$.
Note that, as in FedBuff, we use server momentum, although the theoretical analysis does not include it.
We leave the analysis with momentum for future work.

We model the arrival time and training duration of the clients as in \cite{FedBuff}, modeled after Meta's production FL system, with clients arriving at a constant rate, and with a training duration sampled from a half-normal distribution $Y$, where $Y = |X|$ and $X \sim \mathcal{N}(0,1)$.

For experiments in \cref{fig:Fig3FedBuff}, the server uses learning rates scaled down for staleness, as done in FedBuff and \cite{asyncfl}.
If client $n$ sends an update with staleness $\tau_n$, we scale down its weight by multiplying it by $1/\sqrt{1+\tau_n}$.
To model concurrencies of 100, 500, and 1000 users we vary the constant rate at which clients arrive.
The rates we select to achieve this are 125, 627, and 1253 clients per unit of time, respectively.
This comes from the expected value $\sigma \sqrt{2/\pi}$ of the half-normal $Y$, where $Y = |X|$ and $X \sim \mathcal{N}(0,\sigma^2)$.

For the rest of experiments, we model clients arriving at 100 clients per unit of time, and no weight scaling is performed.

Our implementation is based on the FL Simulator (FLSim), which is a flexible, standalone library written in PyTorch \cite{flsim,pytorch}.

\section{Additional results}
\label{appsec:additional-results}
\begin{figure*}[!ht]
    \centering
    \hspace{2.5cm} \includegraphics[trim={10mm 220mm 10mm 25mm},clip, width=.5\textwidth]{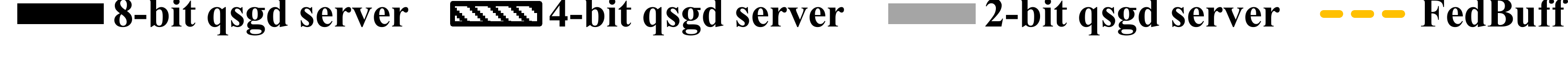}
    \newline
    \includegraphics[trim={10mm 7mm 170mm 5mm},clip, width=.33\textwidth]{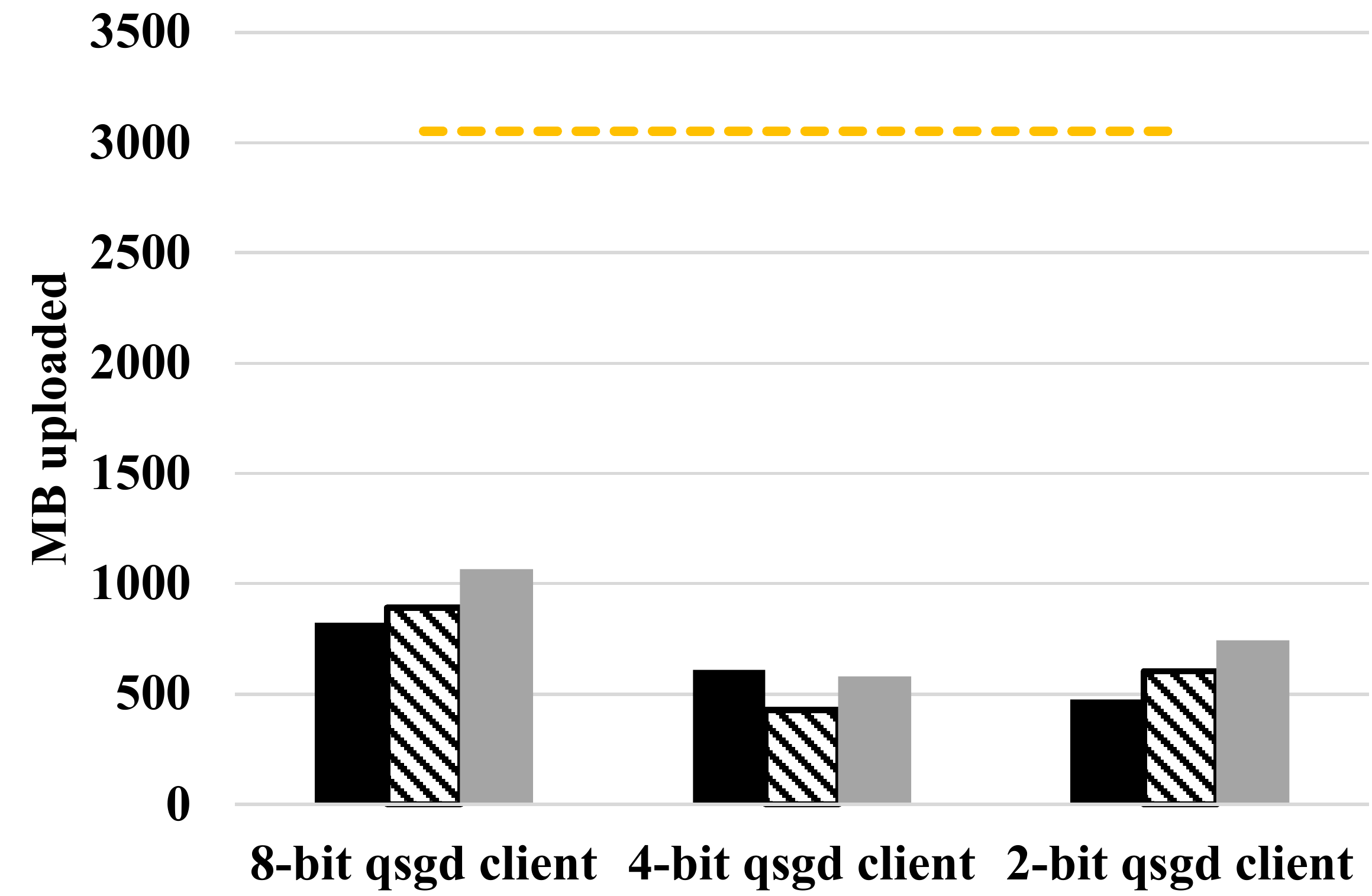}
    \hfill
    \includegraphics[trim={10mm 7mm 170mm 5mm},clip, width=.33\textwidth]{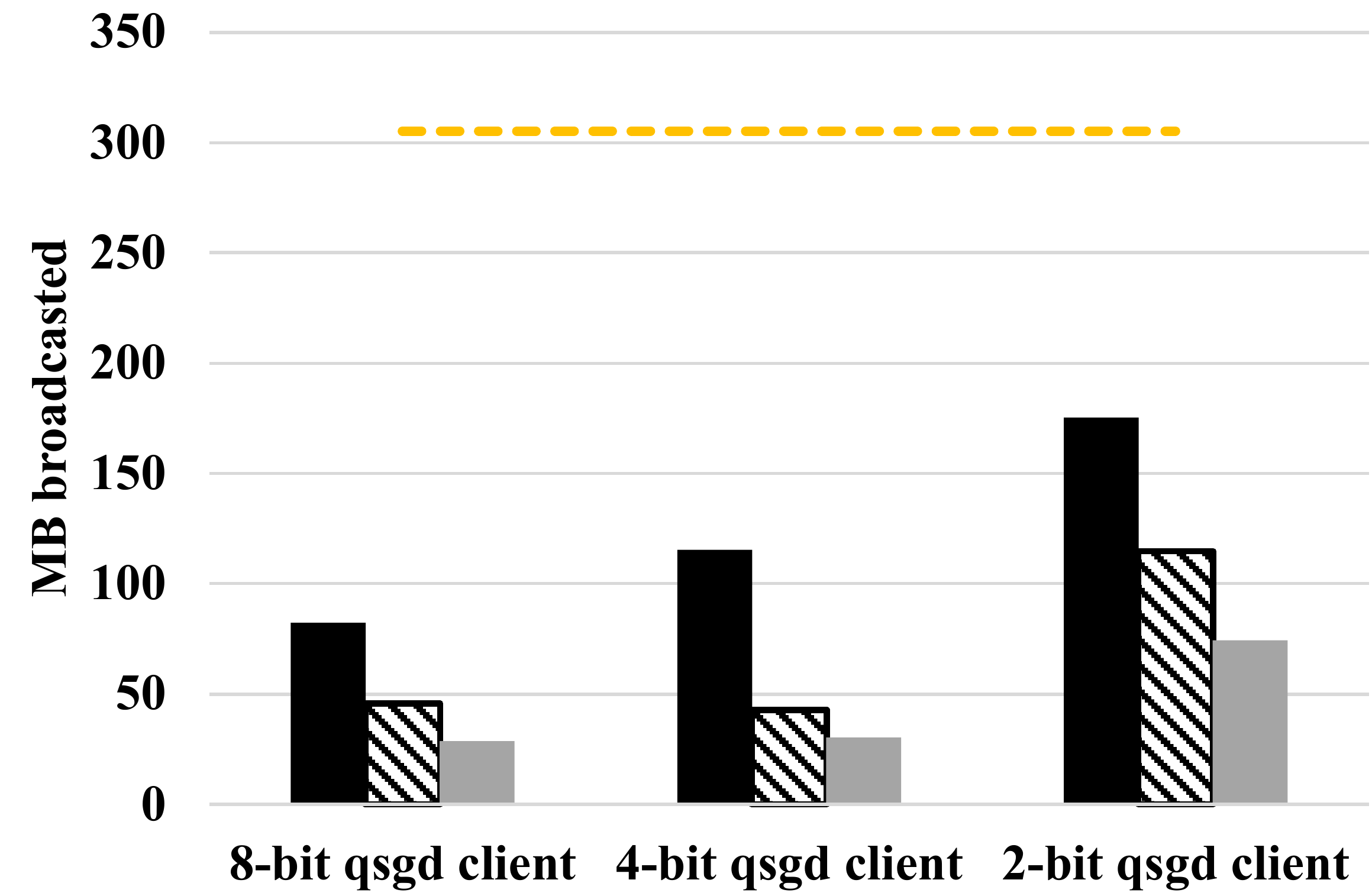}
    \hfill
    \includegraphics[trim={10mm 7mm 170mm 5mm},clip, width=.33\textwidth]{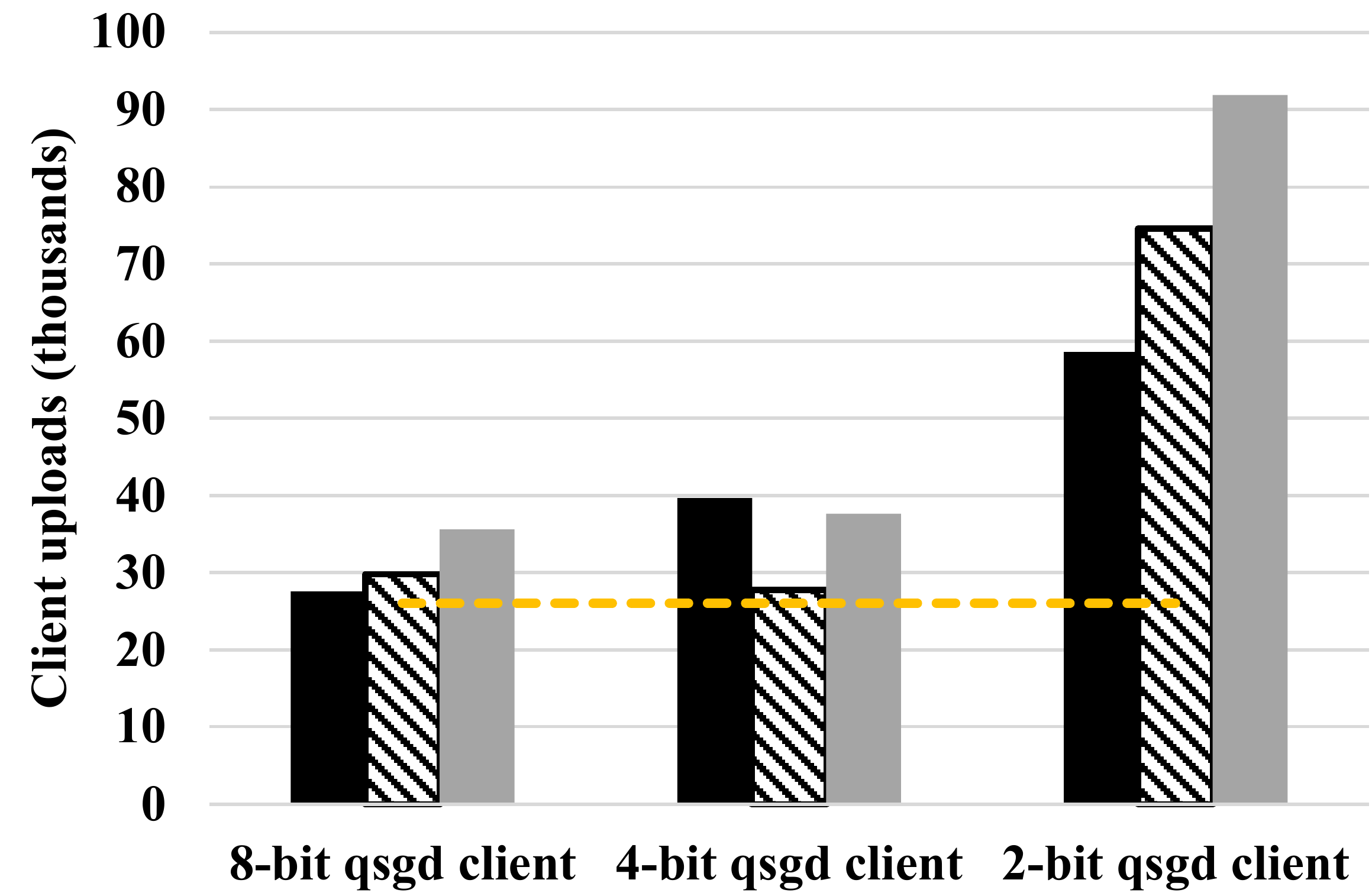}
    \caption{Communication metrics of \algname and FedBuff to reach target validation accuracy (90\%) with varying choices of $\qsgd$ quantizers. In all cases \algname saves at least $3\times$ upload cost and $2\times$ broadcast cost. Notice that in 4-bit $\qsgd$ client and server, the convergence speed is the same for both algorithms, and the upload message size goes down from 3052 MB to 428 MB, a $7\times$ decrease, with analogous decrease in download cost.}
    \label{fig:threefigures}
\end{figure*}

In \cref{fig:threefigures} we observe how, given a client quantizer, quantizing with less precision at the server always results in less total bytes downloaded.
On the other hand, we also observe how quantizing with less precision at the client sometimes results in more total bytes uploaded, e.g., going from 4 to 2-bit $\qsgd$ at the client, with 8-bit $\qsgd$ at the server.
We also see that the number of uploads augments from 4-bit to 2-bit client $\qsgd$, with no significant reduction in total upload bytes.
This illustrates a trade-off between the amount of quantization and the speed of convergence.
In other words, compressing more will ensure fewer bytes per message are sent, but more messages will have to be sent to reach the target accuracy.
However, overall, \algname requires less total number of communication bits compared with FedBuff.
The optimal level of quantization in \algname depends on the choice of the quantizer, as is reflected in our results.

\cref{tab:qsgd_table} shows the performance of \algname with varying levels of $\qsgd$ quantization both on the server and on the client.
This is the data used to produce \cref{fig:threefigures}.
For completeness, we present results with a biased server quantizer in \cref{tab:biased}.
\begin{table}[htbp]
    \caption{Communications metrics of \algname to reach target validation accuracy (90\%), with different combinations of $\qsgd$.}
    \label{tab:qsgd_table}
    \begin{center}
        \begin{tabular}{@{}lllll@{}}
            \toprule
            \multicolumn{2}{l}{Algorithm}                   & Uploads (in thousands) & kB/upload       & kB/download          \\
            \midrule
            \multicolumn{2}{l}{FedBuff}                     & $26.1 \pm 6.7$         & 117.128         & 117.128              \\
            \cmidrule{2-5}
            \multirow{3}{*}{\algname client 8-bit $\qsgd$,} & server 8-bit $\qsgd$   & $27.6 \pm 4.82$ & 29.924      & 29.924 \\
                                                            & server 4-bit $\qsgd$   & $29.8 \pm 2.76$ & 29.924      & 15.380 \\
                                                            & server 2-bit $\qsgd$   & $35.7 \pm 14.5$ & 29.924      & 8.108  \\
            \cmidrule{2-5}
            \multirow{3}{*}{\algname client 4-bit $\qsgd$,} & server 8-bit $\qsgd$   & $39.7 \pm 3.32$ & 15.380      & 29.924 \\
                                                            & server 4-bit $\qsgd$   & $27.8 \pm 10.4$ & 15.380      & 15.380 \\
                                                            & server 2-bit $\qsgd$   & $37.7 \pm 9.10$ & 15.380      & 8.108  \\
            \cmidrule{2-5}
            \multirow{3}{*}{\algname client 2-bit $\qsgd$,} & server 8-bit $\qsgd$   & $58.6 \pm 7.16$ & 8.108       & 29.924 \\
                                                            & server 4-bit $\qsgd$   & $74.6 \pm 35.7$ & 8.108       & 15.380 \\
                                                            & server 2-bit $\qsgd$   & $91.9 \pm 34.4$ & 8.108       & 8.108  \\
            \bottomrule
        \end{tabular}
    \end{center}
\end{table}

\begin{table}[htbp]
    \caption{Communications metrics of \algname to reach target validation accuracy (90\%). \algname-server uses $\topk_k$ to send only the top 10\% of coordinates.}
    \label{tab:biased}
    \begin{center}
        \begin{tabular}{@{}llll@{}}
            \toprule
            Algorithm                       & Uploads (in thousands) & kB/upload & kB/download \\
            \midrule
            FedBuff                         & $26.1 \pm 6.7$         & 117.128   & 117.128     \\
            \midrule
            \algname (client 8-bit $\qsgd$) & $39.5 \pm 10.2$        & 29.924    & 15.404      \\
            \algname (client 4-bit $\qsgd$) & $42.4 \pm 10.4$        & 15.380    & 15.404      \\
            \algname (client 2-bit $\qsgd$) & $94.8 \pm 16.7^*$      & 8.108     & 15.404      \\
            \bottomrule
        \end{tabular}
    \end{center}
    *\footnotesize{The third iteration of 2-bit $\qsgd$ only achieved a maximum of 88.62\% accuracy after 150k client uploads, so it is not used in this computation.}
\end{table}

\section{Convergence analysis and proof}
\label{appsec:convergence-analysis-proof}
In this appendix, we discuss \algname{}'s convergence analysis and provide the main theorem, a sketch of its proof that contains the main ideas, and an detailed proof along with the corollaries we use to back \cref{prop:complexity_order} in the main body.
We also provide the condition on the learning rates necessary for the convergence rate to hold.
For the case of biased server quantizers, we provide a looser bound on the convergence rate.
Finally, we analyze the convergence rate's order of complexity and compare with the rate for FedBuff, i.e., the case with infinite precision.
\cref{prop:complexity_order} follows from this analysis, and it is specifically a direct consequence of \cref{cor:complexity}.

To facilitate the notation, we introduce two quantities, $\alpha_P$ and $\beta_P$, defined as follows:
\begin{equation*}
    \alpha_P := \sum_{p = 0}^{P-1} \eta_\ell^{(p)}, \quad \beta_P := \sum_{p = 0}^{P-1} (\eta_\ell^{(p)})^2.
\end{equation*}

\paragraph{Conditions on the learning rates.}
All learning rates $\eta_g$, $\eta^{(p)}_{\ell}$, for all $p \in \{0, \ldots, P-1\}$, must satisfy
\begin{equation}\label{eq:condition_learning_rates}
    \left(\frac{\alpha_P 3L^2 \eta_g^2}{\delta_s} + L \eta_g \right) \left(1 + \frac{1-\delta_c}{K}\right)P \eta^{(p)}_{\ell} \leq 1.
\end{equation}

For a simpler and less precise bound, if $\eta_g \leq \tfrac{1}{L}$ and $\eta_{\ell}^{(p)} \leq \min \left( \tfrac{K}{2P(K+1-\delta_c)}, \tfrac{\delta_s}{3P} \right)$, the condition is satisfied.

\begin{theorem}[Convergence of \algname{}]
    \label{thm:main-convergence}
    Choosing local learning rates $\eta_{\ell}^{(p)}$ and global learning rate $\eta_g$ that satisfy Condition~\eqref{eq:condition_learning_rates}, and unbiased server and client quantizers, the ergodic convergence rate for $T$ iterations of \algname{} is upper bound by the following:
    \begin{multline}
        \frac{1}{T} \sum_{t=0}^{T-1} \expec{\norm{\nabla f(x^t)}^2} \leq \frac{2\left(f(x^0) - f^* \right)}{\eta_g T \alpha_P} + 3 L^2 \beta_P \left( \eta^2_g \tau_{\max,K}^2 \left( \frac{1-\delta_c}{K \tau_{\max, K}} + 1 \right) + 1 \right)  (\sigma_{\ell}^2 + PG) \\
        {} +  \left(\frac{L \eta_g}{\alpha_P} + 3L^2 \eta_g^2 \sum_{t=1}^{T-1} (1-\delta_s)^t \right) \left(\frac{2-\delta_c}{K}\right)\beta_P \left(\sigma_{\ell}^2 + 4G\right).
    \end{multline}
\end{theorem}

\paragraph{Proof sketch.}
In this proof, we employ $L$-smoothness to bound how the loss function from \eqref{eq:minimization_problem} evolves with our algorithm, i.e.,
\begin{equation} \label{eq:first-decomposition}
    f(x^{t+1}) \leq f(x^t) - \eta_g \left\langle \nabla f(x^t), \overline{\Delta}^t \right\rangle +\frac{L\eta^2_g}{2} \norm{\overline{\Delta}^t}^2.
\end{equation}
Next, we take advantage of the unbiasedness of the client quantizer to compute the expected value of the second term of the RHS of \eqref{eq:first-decomposition}.
Through some further manipulation, we bound the expected value with three terms, each representing a distinct source of error: staleness, quantization, and local drift.
To bound each of these three terms, we utilize the remaining assumptions.
We further bound the $\norm{\overline{\Delta}^t}^2$ term in \eqref{eq:first-decomposition} using an inductive proof.
By rearranging the terms, we are left with some loose terms.
To cancel these out, we derive the condition on the learning rate, ultimately concluding the proof.

A similar proof works for a possibly biased server quantizer, stated in \cref{cor:biased_quantizer}, with the following condition on the learning rates
\begin{equation}\label{eq:condition_learning_rates_biased}
    \left(\frac{12 \alpha_P L^2 \eta_g^2}{\delta_s^2} + L \eta_g \right) \left(1 + \frac{1-\delta_c}{K}\right)P \eta^{(p)}_{\ell} \leq 1.
\end{equation}
Similarly to the unbiased case, for a simpler and less precise bound, if $\eta_g \leq \tfrac{1}{L}$ and $\eta_{\ell}^{(p)} \leq \min \left( \tfrac{K}{2P(K+1-\delta_c)}, \tfrac{\delta_s^2}{12P} \right)$, the condition is satisfied.

\begin{corollary}[Using a biased server quantizer] \label{cor:biased_quantizer}
    Choosing local learning rates $\eta_{\ell}^{(p)}$ and global learning rate $\eta_g$ that satisfy Condition~\eqref{eq:condition_learning_rates_biased}, an unbiased client quantizer, and a \emph{possibly biased} server quantizer, the ergodic convergence rate for $T$ iterations of \algname{} is upper bounded by the following:
    \begin{multline}
        \frac{1}{T} \sum_{t=0}^{T-1} \expec{\norm{\nabla f(x^t)}^2} \leq \frac{2\left(f(x^0) - f^* \right)}{\eta_g T \alpha_P} + 3 L^2 \beta_P \left( \eta^2_g \tau_{\max,K}^2 \left( \frac{1-\delta_c}{K \tau_{\max, K}} + 1 \right) + 1 \right)  (\sigma_{\ell}^2 + PG) \\
        {} +  \left(\frac{L \eta_g}{\alpha_P} +  \frac{12L^2 \eta_g^2}{\delta_s^2} \right) \left(\frac{2-\delta_c}{K}\right)\beta_P \left(\sigma_{\ell}^2 + 4G\right).
    \end{multline}
\end{corollary}

\paragraph{Complexity analysis.}
Under the conditions of \cref{thm:main-convergence}, we can choose learning rates to give the following corollary on complexity.
\begin{corollary}[Complexity order] \label{cor:complexity}
    Let us choose local and global learning rates that satisfy the condition described in \eqref{eq:condition_learning_rates} and define $F^* := f(x^0) - f^*$.
    By also choosing $\eta_\ell = \order{1 / (K\sqrt{TP})}$ and $\eta_g = \order{K}$, for a sufficiently large $T$, it holds that
    \begin{multline} \label{eq:conv_rate}
        \frac{1}{T} \sum_{t=0}^{T-1} \expec{\norm{\nabla f(x^t)}^2} \leq \order{\frac{F^*}{\sqrt{TP}}} + \order{\frac{L^2 (\sigma_{\ell}^2 + PG) \left( \tau_{\max, 1}^2 + \tau_{\max, 1}(1-\delta_c) + 1\right)}{TK^2} } \\
        {} +  \order{\frac{L^2 (2-\delta_c)(\sigma_\ell^2 + 4G)}{\delta_sTK} } + \order{ \frac{L(2-\delta_c)(\sigma_{\ell}^2 + 4G)}{K\sqrt{TP}} } .
    \end{multline}
\end{corollary}
From the bound described in \cref{cor:complexity}, we can see the first standard term for SGD is present in our analysis.
In addition, there are three terms that depend on the maximum staleness $\tau_{\max,1}$, the local variance $\sigma_{\ell}^2$, the gradient bound $G$, and the choice of client and server quantizers.
Observe that the choice of client quantizer affects the order of error in a term that decreases with $1/\sqrt{T}$, while the choice of server quantizer only affects a term that decreases with $1/T$.
This means that although $\sum_{t=1}^{T-1} (1-\delta_s)^t $ can be large, it is bounded by $1/\delta_s$, and the effect of the server quantizer dissipates in time faster than the effect of the client quantizer.
In conclusion, the choice of client quantizer affects the order of error more than the choice of server quantizer.

\subsection{Detailed proof of the main theorem.}
Let us now prove the general version of \cref{thm:main-convergence}, stated here for both biased and un-biased server quantizers, defining $\phi(T)$ as follows:
\begin{equation} \label{eq:phi}
    \phi(T) =
    \begin{cases}
        \frac{2}{\delta_s}\sum_{t=0}^{T-1} (1-\tfrac{\delta_s}{2})^{t-1} \leq \frac{4}{\delta_s^2} & \text{when $Q_s$ is biased,} \\
        \sum_{t=1}^{T-1} (1-\delta_s)^t \leq \frac{1}{\delta_s}                                    & \text{otherwise.}
    \end{cases}
\end{equation}

\begin{theorem}[Convergence of \algname{} - General version]
    Choosing local learning rates $\eta_{\ell}^{(p)}$, and global learning rate $\eta_g$ that satisfy the condition~\eqref{eq:condition_learning_rates}, the ergodic convergence rate for the iterates of \algname{} is upper bound by the following:
    \begin{multline*}
        \frac{1}{T} \sum_{t=0}^{T-1} \expec{\norm{\nabla f(x^t)}^2} \leq \frac{2\left(f(x^0) - f^* \right)}{\eta_g T \alpha_P} + 3 L^2 \beta_P \left( \eta^2_g \tau_{\max,K}^2 \left( \frac{1-\delta_c}{K \tau_{\max, K}} + 1 \right) + 1 \right)  (\sigma_{\ell}^2 + PG) \\
        {} +  \left( 3L^2 \eta_g^2 \phi(T) + \frac{L \eta_g}{\alpha_P} \right) \left(\frac{2-\delta_c}{K}\right)\beta_P \left(\sigma_{\ell}^2 + 4G\right).
    \end{multline*}
\end{theorem}
We use the framework of \cite{FedBuff}, which in turn follows from \cite{perturbedIterate}.
The summary of notation used in the proof is found in \cref{tab:notation-summary}.

\begin{table}[ht]
    \caption{Summary of notation used in the proof.}
    \label{tab:notation-summary}
    \begin{center}
        \begin{tabular}{@{}cl@{}}
            \toprule
            $x^t, \hat x^t$                   & server, shared hidden state at time $t$                           \\
            $L$                               & L-smoothness constant of the loss function                        \\
            $P, p$                            & number, index of local steps at client                            \\
            $K, k$                            & number, index of clients at the buffer                            \\
            $N, n$                            & number, index of total clients                                    \\
            $\eta_g, \eta_\ell^{(p)}$         & server, client (at step $p$) learning rates                       \\
            $Q_s, Q_c$                        & server, client quantizers                                         \\
            $\overline{\Delta}^t, \Delta_k^t$ & server, client $k$'s update at time $t$                           \\
            $\calS^t$                         & set of client indices at the buffer at time $t$                   \\
            $y_{k,p}^t$                       & local state at client $k$, during local step $p$ at time $t$      \\
            $\pm$                             & plus and minus the same quantity, i.e., $a \pm b = a + b - b = a$ \\
            \bottomrule
        \end{tabular}
    \end{center}
\end{table}

Let us start by formally describing the iterates of \cref{alg:server},
\begin{equation*}
    x^{t+1} = x^t + \eta_g \overline{\Delta}^t = x^t + \eta_g \frac{1}{K} Q_c \left( \sum_{k \in \calS^t} \left(- \eta_\ell \sum_{p=1}^P g_k(y_{k,p}^{t - \tau_k(t)}) \right) \right).
\end{equation*}
By \cref{ass:bounded-and-l-smooth-loss-gradient}, we can use $L$-smoothness to compute
\begin{equation}
    \label{eq:l-smoothness-bound}
    \begin{aligned}
        f(x^{t+1}) & \leq f(x^t) - \eta_g \left\langle \nabla f(x^t), \overline{\Delta}^t \right\rangle +\frac{L\eta^2_g}{2} \norm{\overline{\Delta}^t}^2                                                      \\
                   & =   f(x^t) \underbrace{-\frac{\eta_g}{K}\sum_{k\in \calS_t} \left\langle\nabla f(x^t),  \Delta_{k}^{t-\tau_k} \right\rangle}_{T_{1,t}} +\frac{L\eta^2_g}{2} \norm{\overline{\Delta}^t}^2.
    \end{aligned}
\end{equation}
Let us first use \cref{lemma:expected-t1}, which gives us the expected value of $T_{1,t}$,
\begin{equation}
    \begin{aligned}
        \expec{T_{1,t}}
         & = -\frac{\eta_g}{2} \left(\sum_{p=0}^{P-1} \eta^{(p)}_{\ell} \right)\expec{\norm{\nabla f(x^t)}^2}                                                                                                                                              \\
         & + \sum_{p=0}^{P-1} \frac{ \eta_g \eta^{(p)}_{\ell}}{2} \expec{- \norm{\frac{1}{N}\sum_{n = 1}^N \nabla F_n(y^{t-\tau_n}_{n,p})}^2 + \underbrace{\norm{\nabla f(x^t) -  \frac{1}{N}\sum_{n = 1}^N \nabla F_n(y^{t-\tau_n}_{n,p})}^2}_{T_{2,t}}}.
    \end{aligned}
    \label{eq:T1}
\end{equation}
Using the definition of $\alpha_P$, we can plug \eqref{eq:T1} into \eqref{eq:l-smoothness-bound}, and take expectations to obtain
\begin{equation}
    \begin{aligned}
        \expec{f(x^{t+1})} & \leq \expec{f(x^t)}  -\frac{\eta_g}{2} \alpha_P \expec{\norm{\nabla f(x^t)}^2} -\frac{\eta_g}{2}\sum_{p=0}^{P-1}\eta_{\ell}^{(p)} \expec{ \norm{\frac{1}{N}\sum_{n = 1}^N \nabla F_n(y^{t-\tau_n}_{n,p})}^2} \\
                           & +\frac{\eta_g}{2}\alpha_P\expec{T_{2,t}} + \frac{L\eta^2_g}{2}\expec{\norm{ \overline{\Delta}^t }^2}.
    \end{aligned}
\end{equation}

If we rearrange and sum for $t = 0, \ldots, T - 1$, we obtain
\begin{equation}
    \begin{aligned}
        \frac{2}{\eta_g}\expec{f(x^{T}) - f(x^0)} & \leq -\alpha_P \sum_{t=0}^{T-1} \expec{ \norm{\nabla f(x^t)}^2 } - \sum_{t=0}^{T-1} \sum_{p=0}^{P-1}\eta_{\ell}^{(p)} \expec{\norm{\frac{1}{N}\sum_{n = 1}^N \nabla F_n(y^{t-\tau_n}_{n,p})}^2} \\
                                                  & + \alpha_P \sum_{t=0}^{T-1} \expec{T_{2,t}} + L\eta_g \sum_{t=0}^{T-1} \expec{\norm{ \overline{\Delta}^t }^2}.
    \end{aligned}
\end{equation}
Dividing by $T$ and rearranging further, we obtain
\begin{equation}
    \label{eq:rearranged}
    \begin{aligned}
        \frac{1}{T} \sum_{t=0}^{T-1} \expec{\norm{\nabla f(x^t)}^2} & \leq \frac{2}{\eta_g T \alpha_P}\expec{f(x^{0}) - f(x^T)}  - \frac{1}{T\alpha_P} \sum_{t=0}^{T-1} \sum_{p=0}^{P-1}\eta_{\ell}^{(p)} \expec{\norm{\frac{1}{N}\sum_{n = 1}^N \nabla F_n(y^{t-\tau_n}_{n,p})}^2} \\
                                                                    & + \frac{1}{T} \sum_{t=0}^{T-1} \expec{T_{2,t}} + \frac{L\eta_g}{T\alpha_P} \sum_{t=0}^{T-1} \expec{\norm{ \overline{\Delta}^t }^2}.
    \end{aligned}
\end{equation}
Notice that the LHS of our expression is now identical to the problem statement.
Let us continue to develop the RHS.
Using \cref{lemma:expected-t2}, we can bound $\sum_{t=0}^{T-1} \expec{T_{2,t}}$ with
\begin{equation}
    \sum_{t=0}^{T-1} \expec{T_{2,t}} \leq 3 L^2 T \beta_P \left( \eta^2_g \tau_{\max,K}^2 \left (\frac{1-\delta_c}{K \tau_{\max, K}} + 1 \right) + 1 \right)  (\sigma_{\ell}^2 + PG) + 3L^2 \eta_g^2 \phi(T) \sum_{t=0}^{T-2} \expec{\norm{\overline{\Delta}^t}^2},
\end{equation}
where $\phi(T)$ is a function that depends on the choice of quantizer at the server, as well as the number of iterations $T$, see \eqref{eq:phi}.

Plugging this bound into \eqref{eq:rearranged}, we obtain
\begin{equation}
    \begin{aligned}
        \frac{1}{T} \sum_{t=0}^{T-1} \expec{\norm{\nabla f(x^t)}^2} & \leq \frac{2}{\eta_g T \alpha_P}\expec{f(x^{0}) - f(x^T)}  - \frac{1}{T\alpha_P} \sum_{t=0}^{T-1} \sum_{p=0}^{P-1}\eta_{\ell}^{(p)} \expec{\norm{\frac{1}{N}\sum_{n = 1}^N \nabla F_n(y^{t-\tau_n}_{n,p})}^2} \\
                                                                    & + 3 L^2 \beta_P \left( \eta^2_g \tau_{\max,K}^2 \left (\frac{1-\delta_c}{K \tau_{\max, K}} + 1 \right) + 1 \right)  (\sigma_{\ell}^2 + PG)                                                                    \\
                                                                    & +  \left( 3L^2 \eta_g^2 \phi(T) + \frac{L \eta_g}{\alpha_P} \right) \frac{1}{T} \sum_{t=0}^{T-1} \expec{\norm{\overline{\Delta}^t}}^2.
    \end{aligned}
\end{equation}
Now, we can bound the last term of the RHS with the bound derived in \cref{sec:bounding-server-step}, which states
\begin{equation}
    \expec{\norm{\overline{\Delta}^t }^2} \leq \left(\frac{2-\delta_c}{K}\right)\beta_P \left(\sigma_{\ell}^2 + 4G\right) + \left(1 + \frac{1-\delta_c}{K}\right)P  \sum_{p=0}^{P-1}   (\eta^{(p)}_{\ell})^2 \expec{\norm{\frac{1}{N} \sum_{n = 1}^N \nabla F_n(y^{t-\tau_n}_{n,p}) }^2},
\end{equation}
and we obtain
\begin{equation}
    \label{eq:long-expansion}
    \begin{aligned}
        \frac{1}{T} \sum_{t=0}^{T-1} \expec{\norm{\nabla f(x^t)}^2} & \leq \frac{2}{\eta_g T \alpha_P}\expec{f(x^{0}) - f(x^T)}                                                                                                                                                                                                   \\
                                                                    & - \frac{1}{T\alpha_P} \sum_{t=0}^{T-1} \sum_{p=0}^{P-1}\eta_{\ell}^{(p)} \expec{\norm{\frac{1}{N}\sum_{n = 1}^N \nabla F_n(y^{t-\tau_n}_{n,p})}^2}                                                                                                          \\
                                                                    & + 3 L^2 \beta_P \left( \eta^2_g \tau_{\max,K}^2 \left (\frac{1-\delta_c}{K \tau_{\max, K}} + 1 \right) + 1\right)  (\sigma_{\ell}^2 + PG)                                                                                                                   \\
                                                                    & +  \left( 3L^2 \eta_g^2 \phi(T) + \frac{L \eta_g}{\alpha_P} \right) \left(\frac{2-\delta_c}{K}\right)\beta_P \left(\sigma_{\ell}^2 + 4G\right)                                                                                                              \\
                                                                    & + \left( 3L^2 \eta_g^2 \phi(T) + \frac{L \eta_g}{\alpha_P} \right) \left(1 + \frac{1-\delta_c}{K}\right)\frac{P}{T}\sum_{t=0}^{T-1}   \sum_{p=0}^{P-1}   (\eta^{(p)}_{\ell})^2 \expec{\norm{\frac{1}{N} \sum_{n = 1}^N \nabla F_n(y^{t-\tau_n}_{n,p}) }^2}.
    \end{aligned}
\end{equation}
Observe that we can impose that the sum of the second term and the last term of the RHS be non-positive, as long as
\begin{equation}
    \left( 3L^2 \eta_g^2 \phi(T) + \frac{L \eta_g}{\alpha_P} \right) \left(1 + \frac{1-\delta_c}{K}\right)P \eta^{(p)}_{\ell} - \frac{1}{\alpha_P} \leq 0, \quad \forall p \in \{0, \ldots, P-1\}.
\end{equation}
Rearranging, we achieve the equivalent condition
\begin{equation}
    \label{eq:condition}
    \left(\alpha_P 3L^2 \eta_g^2 \phi(T) + L \eta_g \right) \left(1 + \frac{1-\delta_c}{K}\right)P \eta^{(p)}_{\ell} \leq 1, \quad \forall p \in \{0, \ldots, P-1\}.
\end{equation}
For a simpler, less precise bound, remark that if $\eta_g \leq \tfrac{1}{L}$ and $\eta_{\ell}^{(p)} \leq \min \left( \tfrac{K}{2P(K+1-\delta_c)}, \tfrac{1}{3\phi(T)P} \right)$ the condition is satisfied.
Assuming that \eqref{eq:condition} is satisfied, \eqref{eq:long-expansion} becomes
\begin{equation}
    \begin{aligned}
        \frac{1}{T} \sum_{t=0}^{T-1} \expec{\norm{\nabla f(x^t)}^2} & \leq \frac{2}{\eta_g T \alpha_P}\expec{f(x^{0}) - f(x^T)} + 3 L^2 \beta_P \left( \eta^2_g \tau_{\max,K}^2 \left (\frac{1-\delta_c}{K \tau_{\max, K}} + 1 \right) + 1\right)  (\sigma_{\ell}^2 + PG) \\
                                                                    & +  \left( 3L^2 \eta_g^2 \phi(T) + \frac{L \eta_g}{\alpha_P} \right) \left(\frac{2-\delta_c}{K}\right)\beta_P \left(\sigma_{\ell}^2 + 4G\right).
    \end{aligned}
\end{equation}

Finally, using $f^*$ as a lower bound for all values of $f$, we conclude the proof.

\subsection{Technical lemmas}
\begin{lemma}[Stochastic gradient bound] \label{lemma:stochastic-gradient-bound}
    For any $x\in \mathbb{R}^d$, and any learning rates $\eta_{\ell}^{(p)}$,
    \begin{equation}
        \expec{\norm{\sum_{p=0}^{P-1} \eta_{\ell}^{(p)} g_k(x)}^2} \leq \beta_P\left( \sigma_{\ell}^2 + PG\right).
    \end{equation}
\end{lemma}
\begin{proof}
    Adding and subtracting the true client gradient,
    \begin{equation}
        \expec{\norm{\sum_{p=0}^{P-1} \eta_{\ell}^{(p)} (g_k(x) \pm F_k(x))}^2} = \sum_{p=0}^{P-1} \expec{\norm{\eta_{\ell}^{(p)}(g_k(x) - F_k(x))}^2} + \expec{\norm{\sum_{p=0}^{P-1} \eta_{\ell}^{(p)} F_k(x)}^2}.
    \end{equation}
    where the equality follows from the unbiasedness of $g_k$.
    Applying Cauchy-Schwarz gives us the statement of the lemma.
\end{proof}

\begin{lemma} [Sum of unbiasedly quantized terms bound] \label{lemma:sum-unbiasedly-quantized-bound}
    For any set of $n \geq 1$ vectors $\{x_n \in \mathbb{R}^d, n = 1,\ldots,n\}$, and any unbiased compression operator $Q: \mathbb{R}^d \to \mathbb{R}^d$ satisfying \cref{def:quantization}, we can ensure
    \begin{equation*}
        \mathbb{E}_Q \norm{\sum_{n = 1}^n Q(x_n)}^2 \leq \norm{\sum_{n = 1}^n x_n}^2 + (1-\delta) \sum_{n = 1}^n \norm{x_n}^2.
    \end{equation*}
\end{lemma}
\begin{proof}
    From the identity
    \begin{equation*}
        \norm{a + b}^2 = \norm{a}^2 +  \norm{b}^2 + 2\langle a, b \rangle,
    \end{equation*}
    we can deduce
    \begin{multline}
        \mathbb{E}_Q \norm{\sum_{n = 1}^n Q(x_n)}^2 = \mathbb{E}_Q \norm{\sum_{n = 1}^n \left( Q(x_n) \pm x_n\right)}^2 = \mathbb{E}_Q \norm{\sum_{n = 1}^n \left( Q(x_n) - x_n\right)}^2 + \norm{\sum_{n = 1}^n x_n}^2 \\
        {}+ 2\mathbb{E}_Q \langle \sum_{n = 1}^n \left( Q(x_n) - x_n\right), \sum_{n = 1}^n x_n\rangle =  \mathbb{E}_Q \norm{\sum_{n = 1}^n \left (Q(x_n) - x_n\right)}^2 + \norm{\sum_{n = 1}^n x_n}^2,
    \end{multline}
    where the last equality follows from the unbiasedness of $Q$.
    Similarly, since $Q(x_n) - x_n$ are independent with respect to the randomness of $Q$, and $Q$ is unbiased,
    \begin{equation}
        \mathbb{E}_Q \norm{\sum_{n = 1}^n \left (Q(x_n) - x_n\right)}^2 =  \mathbb{E}_Q \sum_{n = 1}^n \norm{ Q(x_n) - x_n }^2 \leq (1-\delta) \sum_{n = 1}^n \norm{x_n}^2.
    \end{equation}
    Joining the past two expressions gives us the statement of the lemma.
\end{proof}

\begin{lemma}[Expected value of $T_{1,t}$] \label{lemma:expected-t1}
    For the iterates of \algname{}, and defining $T_{1,t}$ as
    $$T_{1,t} := -\frac{\eta_g}{K}\sum_{k\in \calS_t} \left\langle\nabla f(x^t),  \Delta_{k}^{t-\tau_k} \right\rangle,$$
    it holds that
    \begin{equation}
        \begin{aligned}
            \expec{T_{1,t}}
             & = -\frac{\eta_g}{2} \left(\sum_{p=0}^{P-1} \eta^{(p)}_{\ell} \right)\expec{\norm{\nabla f(x^t)}^2}                                                                                                                                             \\
             & + \sum_{p=0}^{P-1} \frac{ \eta_g \eta^{(p)}_{\ell}}{2} \expec{- \norm{\frac{1}{N}\sum_{n = 1}^N \nabla F_n(y^{t-\tau_n}_{n,p})}^2 + \underbrace{\norm{\nabla f(x^t) - \frac{1}{N}\sum_{n = 1}^N \nabla F_n(y^{t-\tau_n}_{n,p})}^2}_{T_{2,t}}}.
        \end{aligned}
    \end{equation}
\end{lemma}
\begin{proof}
    First, let us use the definition of $\Delta_{k}^{t-\tau_k}$, which is the client $k$'s update sent to the server at time $t-\tau_k$. We can take the expectation of $T_{1,t}$ to obtain
    \begin{equation}
        \expec{T_{1,t}}
        = \expec{-\frac{\eta_g}{K} \sum_{k\in \calS_t}  \left\langle \nabla f(x^t), Q_c \left( \sum_{p=0}^{P-1} \eta^{(p)}_{\ell} g_k(y^{t-\tau_k}_{k,p}) \right) \right\rangle}
        = \expec{-\frac{\eta_g}{K} \sum_{k\in \calS_t}  \sum_{p=0}^{P-1} \eta^{(p)}_{\ell} \left\langle \nabla f(x^t),   g_k(y^{t-\tau_k}_{k,p}) \right\rangle },
    \end{equation}
    because the client quantizer $Q_c$ is unbiased, and its internal randomness is independent of all other variables.
    Now, following the same logic as in \cite{FedBuff} we use the conditional expectation
    \begin{equation*}
        \expec{\cdot}:= \mathbb{E}_{\mathcal{H}} \mathbb{E}_{n \sim [N]} \mathbb{E}_{g_n \mid n, \mathcal{H}}[\cdot],
    \end{equation*}
    where $\mathcal{H}$ corresponds to the randomness of the history of the iterates, and $g_n \mid n, \mathcal{H}$ corresponds to the randomness of the stochastic gradient $g_n$, conditioned to the fact that we chose client $n$, and we have the iterates defined by $\mathcal{H}$.
    We have used $n \sim [N]$ to indicate that the client $n$ is sampled randomly and uniformly over $1, \ldots, N$ at each time $t$.
    The expectation with respect to the quantization randomness is implicit in this notation.
    Using \cref{ass:unbiased-stochastic-gradient}, we can ensure that the stochastic gradient is unbiased, and we obtain
    \begin{equation}
        \begin{aligned}
            \expec{T_{1,t}} = & - \expec{ \frac{\eta_g}{K}  \sum_{k\in \calS_t} \sum_{p=0}^{P-1} \eta^{(p)}_{\ell}  \left\langle \nabla f(x^t),   g_k(y^{t-\tau_k}_{k,p}) \right\rangle }                                                              \\
            =                 & - \eta_g \mathbb{E}_{\mathcal{H}} \left[ \frac{1}{N} \sum_{n = 1}^N \sum_{p=0}^{P-1} \eta^{(p)}_{\ell}  \mathbb{E}_{g_n \mid n \sim[N]}  \left\langle \nabla f(x^t),    g_n(y^{t-\tau_n}_{n,p})  \right\rangle \right] \\
            =                 & - \eta_g \mathbb{E}_{\mathcal{H}} \left[ \sum_{p=0}^{P-1} \eta^{(p)}_{\ell} \left\langle \nabla f(x^t), \frac{1}{N} \sum_{n = 1}^N \nabla F_n(y^{t-\tau_n}_{n,p})  \right\rangle \right].
        \end{aligned}
    \end{equation}
    Notice that we have removed the dependency on the client selection, so we can simply write
    \begin{equation}
        \expec{T_{1,t}} = - \eta_g \expec{\sum_{p=0}^{P-1} \eta^{(p)}_{\ell}  \left\langle \nabla f(x^t), \frac{1}{N}\sum_{n = 1}^N \nabla F_n(y^{t-\tau_n}_{n,p}) \right\rangle}.
    \end{equation}
    Finally, from the identity
    \begin{equation*}
        \langle a, b \rangle = \frac{1}{2}(\norm{a}^2 + \norm{b}^2 - \norm{a-b}^2),
    \end{equation*}
    we can develop the previous inner product and obtain
    \begin{equation}
        \begin{aligned}
            \expec{T_{1,t}}
             & = -\frac{\eta_g}{2} \left(\sum_{p=0}^{P-1} \eta^{(p)}_{\ell} \right)\expec{\norm{\nabla f(x^t)}^2}                                                                                                                      \\
             & + \sum_{p=0}^{P-1} \frac{ \eta_g \eta^{(p)}_{\ell}}{2} \expec{- \norm{\frac{1}{N}\sum_{n = 1}^N \nabla F_n(y^{t-\tau_n}_{n,p})}^2 + \norm{\nabla f(x^t) - \frac{1}{N}\sum_{n = 1}^N \nabla F_n(y^{t-\tau_n}_{n,p})}^2}.
        \end{aligned}
    \end{equation}
\end{proof}

\begin{lemma}[Expected value of $T_{2,t}$]
    \label{lemma:expected-t2}
    For the iterates of \algname{}, and defining $T_{2,t}$ as
    \begin{equation*}
        T_{2,t} := \norm{\nabla f(x^t) - \frac{1}{N}\sum_{n = 1}^N \nabla F_n(y^{t-\tau_n}_{n,p})}^2,
    \end{equation*}
    it holds that, for any $T > 0$,
    \begin{equation}
        \sum_{t=0}^{T-1} \expec{T_{2,t}} \leq 3 L^2 T \beta_P \left( \eta^2_g \tau_{\max,K}^2 \left (\frac{1-\delta_c}{K \tau_{\max, K}} + 1 \right) + 1\right)  (\sigma_{\ell}^2 + PG) + 3L^2 \eta_g^2 \phi(T) \sum_{t=0}^{T-2} \expec{\norm{\overline{\Delta}^t}^2}.
    \end{equation}
\end{lemma}
\begin{proof}
    Let us start expanding the expected value of $T_{2,t}$, and applying the Cauchy–Schwarz inequality
    \begin{equation}
        \expec {T_{2,t}} = \expec{\norm{ \frac{1}{N}\sum_{n = 1}^N  \nabla F_n(x^t) - \frac{1}{N} \sum_{n = 1}^N \nabla F_n(y^{t-\tau_n}_{n,p})}^2} \leq \frac{1}{N} \sum_{n = 1}^N \expec{\norm{\nabla F_n(x^t) - \nabla F_n(y^{t-\tau_n}_{n,p})}^2}.
    \end{equation}
    We apply the $L$-smoothness assumption from \cref{ass:bounded-and-l-smooth-loss-gradient} to each term in the previous sum, and telescope them with
    \begin{equation}
        \expec{\norm{\nabla F_n(x^t) - \nabla F_n(y^{t-\tau_n}_{n,p})}^2} \leq L^2 \expec{\norm{x^t -y^{t-\tau_n}_{n,p}}^2} = L^2 \expec{\norm{x^t \pm x^{t - \tau_n} \pm \hat x^{t - \tau_n} -y^{t-\tau_n}_{n,p}}^2}.
    \end{equation}
    Again, by Cauchy–Schwarz, we obtain the following decomposition
    \begin{equation}
        \expec {T_{2,t}} \leq \frac{3L^2}{N}\sum_{n = 1}^N \expec{ \underbrace{\norm{x^t - x^{t-\tau_n} }^2}_{\text{staleness}} + \underbrace{\norm{x^{t-\tau_n} - \hat x^{t-\tau_n} }^2}_{\text{quantization}} + \underbrace{\norm{\hat x^{t-\tau_n} - y^{t-\tau_n}_{n,p} }^2}_{\text{local drift}} }.
    \end{equation}
    In \cref{sec:bounding-staleness-quantization-drift}, we derive the bounds for these three terms.
    We can use the three derived bounds in \eqref{eq:bound-staleness}, \eqref{eq:bound-drift} and \eqref{eq:bound-quantization-term}, to sum for $t=0, \ldots, T-1$ and obtain
    \begin{equation}
        \sum_{t=0}^{T-1} \expec{T_{2,t}} \leq 3 L^2 T \beta_P \left(\eta^2_g \tau_{\max,K}^2   (\sigma_{\ell}^2 + PG) \left (\frac{1-\delta_c}{K \tau_{\max, K}} + 1 \right) + ( \sigma_{\ell}^2 + PG ) \right) + 3L^2 \eta_g^2 \phi(T) \sum_{t=0}^{T-2} \expec{\norm{\overline{\Delta}^t}^2}.
    \end{equation}
\end{proof}

\subsection{Bounding the effects of staleness, quantization, and local drift}
\label{sec:bounding-staleness-quantization-drift}

\paragraph{Staleness.}
The staleness term gives us the error inferred by the $\tau_n$ steps of difference between the current server model and the model that the client is training with at the time it sends the update.
We can start by using the definition of $x^t$,
\begin{equation}
    \begin{aligned}
        \norm{x^t - x^{t-\tau_n}}^2 = \norm{\sum^{t-1}_{\rho=t-\tau_n} \overline{\Delta}^\rho }^2 = \norm{\sum^{t-1}_{\rho=t-\tau_n} \frac{\eta_g}{K} \sum_{j_\rho \in \calS_{\rho}} \Delta_{j_\rho}^{\rho}}^2 = \frac{\eta^2_g}{K^2}\norm{\sum^{t-1}_{\rho=t-\tau_n} \sum_{j_\rho \in \calS_{\rho}} Q_c \left( \sum_{p'=0}^{P-1} \eta^{(p')}_{\ell}  g_{j_\rho}(y^{\rho}_{j_\rho, p'}) \right)}^2.
    \end{aligned}
\end{equation}
We can then use the unbiasedness of $Q_c$ and apply \cref{lemma:sum-unbiasedly-quantized-bound}, so
\begin{equation}
    \begin{aligned}
        \expec{\norm{x^t - x^{t-\tau_n}}^2} & \leq \frac{\eta^2_g}{K^2}\expec{\norm{\sum^{t-1}_{\rho=t-\tau_n} \sum_{j_\rho \in \calS_{\rho}} \sum_{p'=0}^{P-1} \eta^{(p')}_{\ell}  g_{j_\rho}(y^{\rho}_{j_\rho, p'}) }^2}             \\
                                            & + (1-\delta_c)\frac{\eta^2_g}{K^2} \expec{\sum^{t-1}_{\rho=t-\tau_n} \sum_{j_\rho \in \calS_{\rho}} \norm{ \sum_{p'=0}^{P-1} \eta^{(p')}_{\ell}  g_{j_\rho}(y^{\rho}_{j_\rho, p'}) }^2}.
    \end{aligned}
\end{equation}
We can now apply Cauchy-Schwarz, the definition of $\beta_P$, and \cref{lemma:stochastic-gradient-bound} to obtain
\begin{equation} \label{eq:bound-staleness}
    \begin{aligned}
        \expec{\norm{x^t - x^{t-\tau_n}}^2} & \leq \frac{\eta^2_g}{K^2}\tau_{\max,K}^2K^2 \beta_P (\sigma_{\ell}^2 + PG ) + (1-\delta_c)\frac{\eta^2_g}{K^2} \tau_{\max,K} K \beta_P  (\sigma_{\ell}^2 + PG) \\
                                            & = \eta^2_g \tau_{\max,K}^2 \beta_P  (\sigma_{\ell}^2 + PG) \left (\frac{1-\delta_c}{K \tau_{\max, K}} + 1 \right).
    \end{aligned}
\end{equation}

\paragraph{Local drift.}
The local drift term tells us the norm of the difference between the $p$-th local step and the initial hidden state with which the local training started.
\begin{equation}
    \expec{\norm{\hat x^{t-\tau_n} - y^{t-\tau_n}_{n,p}}^2 } = \expec{\norm{y^{t-\tau_n}_{i,0} - y^{t-\tau_n}_{n,p}}^2} = \expec{ \norm{\sum_{p'=0}^{p-1}  \eta^{(p')}_{\ell}g_n(y^{t-\tau_n}_{n,p'}) }^2}.
\end{equation}
Applying the definition of $\beta_P$ and \cref{lemma:stochastic-gradient-bound} yields
\begin{equation} \label{eq:bound-drift}
    \expec{\norm{\hat x^{t-\tau_n} - y^{t-\tau_n}_{n,p}}^2 } \leq \beta_P( \sigma_{\ell}^2 + PG ).
\end{equation}

\paragraph{Quantization.}
The quantization term tells us how much the server state and the hidden state differ at time $t-\tau_n$.
This term requires more work, and we divide it into two cases, depending on whether the server quantizer $Q_s$ is biased or not.
Both these cases are discussed in \cref{sec:quantization-lemmas}.
We then add the expectation of the quantization terms for $t=0, \ldots, T-1$, as this simplifies the notation with the definition of $\phi(T)$ we introduced in \eqref{eq:phi}.
For clarity, we shall restate it here:
\begin{equation}
    \phi(T) =
    \begin{cases}
        \frac{2}{\delta_s}\sum_{t=0}^{T-1} (1-\tfrac{\delta_s}{2})^{t-1} \leq \frac{4}{\delta_s^2} & \text{when $Q_s$ is biased,} \\
        \sum_{t=1}^{T-1} (1-\delta_s)^t \leq \frac{1}{\delta_s}                                    & \text{otherwise.}
    \end{cases}
\end{equation}
The bounds made in both cases are simple geometric sums.
This leads us to the following bound on the quantization term
\begin{equation} \label{eq:bound-quantization-term}
    \sum_{t=0}^{T-1} \expec{\norm{x^{t-\tau_n} - \hat x^{t-\tau_n} }^2} \leq \eta_g^2 \phi(T) \sum_{t=0}^{T-2} \expec{\norm{\overline{\Delta}^t}^2}.
\end{equation}
The unbiased case is a direct consequence of \cref{cor:sum-unbiased-terms}, and the biased case is proven in \cref{cor:sum-biased-terms}.
As we can see, an unbiased server quantizer can help us prove a better bound on convergence.

\subsection{Quantization lemmas}
\label{sec:quantization-lemmas}

\begin{lemma} [Hidden state bound with an unbiased quantizer] \label{lemma:hidden-state-unbiased}
    Given an unbiased quantizer $Q_s$, for the iterations of \algname{}, it holds that
    \begin{equation}
        \expec{\norm{x^{t} - \hat x^{t}}^2} \leq \eta_g^2 \sum_{s=0}^{t-1} (1-\delta_s)^{t-s} \expec{\norm{\overline{\Delta}^s}^2}.
    \end{equation}
\end{lemma}
\begin{proof}
    We prove the statement by induction on $t$.
    For $t=0$, $x^t = \hat x^t$, so the statement holds.
    We assume the lemma is true for $t$, and prove for $t+1$.
    First, by definition of $\hat x^{t+1}$,
    \begin{equation} \label{eq:unroll-diff}
        \expec{\norm{x^{t+1} - \hat x^{t+1}}^2} = \expec{\norm{x^{t+1} - \hat x^{t} - Q_s(x^{t+1} - \hat x^t)}^2} \leq (1-\delta_s) \expec{\norm{x^{t+1} - \hat x^t}^2}.
    \end{equation}
    Then, by the definition of $x^{t+1}$,
    \begin{equation} \label{eq:using-unbiased}
        \expec{\norm{x^{t+1} - \hat x^t}^2} = \expec{\norm{x^{t} + \eta_g \overline{\Delta}^t - \hat x^t}^2} = \expec{\norm{x^{t} - \hat x^{t}}^2} + \eta_g^2 \expec{\norm{\overline{\Delta}^t}^2},
    \end{equation}
    where the last equality follows from the unbiasedness of $Q_s$.
    Therefore, plugging \eqref{eq:using-unbiased} into \eqref{eq:unroll-diff}, we obtain
    \begin{equation}
        \expec{\norm{x^{t+1} - \hat x^{t+1}}^2} \leq (1-\delta_s) \left( \expec{\norm{x^{t} - \hat x^{t}}^2} + \eta_g^2 \expec{\norm{\overline{\Delta}^t}^2} \right),
    \end{equation}
    and using induction to substitute the first term concludes the proof,
    \begin{equation}
        \begin{aligned}
            \expec{\norm{x^{t+1} - \hat x^{t+1}}^2} & \leq (1-\delta_s) \left(\eta_g^2 \expec{\norm{\overline{\Delta}^t}^2} + \eta_g^2 \sum_{s=0}^{t-1} (1-\delta_s)^{t-s} \expec{\norm{\overline{\Delta}^s}^2}\right) \\
                                                    & = \eta_g^2 \sum_{s=0}^{t} (1-\delta_s)^{t+1-s} \expec{\norm{\overline{\Delta}^s}^2}.
        \end{aligned}
    \end{equation}
\end{proof}

\begin{corollary} [Sum of terms from \cref{lemma:hidden-state-unbiased}] \label{cor:sum-unbiased-terms}
    From \cref{lemma:hidden-state-unbiased}, summing for $t = 0, \ldots, T-1$, it follows that
    \begin{equation}
        \sum_{t=0}^{T-1} \expec{\norm{x^{t} - \hat x^{t}}^2} \leq \eta_g^2 \sum_{s=1}^{T-1} (1-\delta_s)^s \sum_{t=0}^{T-2} \expec{\norm{\overline{\Delta}^t}^2}.
    \end{equation}
\end{corollary}
\begin{proof}
    \begin{equation}
        \begin{aligned}
            \sum_{t=0}^{T-1} \expec{\norm{x^{t} - \hat x^{t}}^2} & \leq \sum_{t=0}^{T-1} \sum_{s=0}^{t-1} (1-\delta_s)^{t-s} \eta_g^2 \expec{ \norm{\overline{\Delta}^s}^2} = \sum_{s=0}^{T-2} \sum_{t=s+1}^{T-1} (1-\delta_s)^{t-s} \eta_g^2 \expec{\norm{\overline{\Delta}^s}^2} \\
                                                                 & = \eta_g^2 \sum_{s=0}^{T-2} \expec{ \norm{\overline{\Delta}^s}^2} \sum_{t=s+1}^{T-1} (1-\delta_s)^{t-s} \leq \eta_g^2\sum_{s=0}^{T-2} \expec{ \norm{\overline{\Delta}^s}^2} \sum_{t=1}^{T-1} (1-\delta_s)^t .
        \end{aligned}
    \end{equation}
\end{proof}

\begin{lemma} [Hidden state bound with a general quantizer] \label{lemma:hidden-state-biased}
    Given any quantizer $Q_s$, for the iterations of \algname{}, it holds that
    \begin{equation}
        \expec{\norm{x^{t} - \hat x^{t}}^2} \leq \eta_g^2 \frac{2}{\delta_s} \sum_{s=0}^{t-1} \left(1- \frac{\delta_s}{2 } \right)^{t-1-s} \expec{ \norm{\overline{\Delta}^s}^2}.
    \end{equation}
\end{lemma}
\begin{proof}
    The proof follows in a similar way to the unbiased case, by induction on $t$.
    For $t=0$, $x^t = \hat x^t$, so the statement holds.
    We assume that the lemma is true for $t$, and prove for $t+1$.
    First, by definition of $\hat x^{t+1}$,
    \begin{equation}
        \expec{\norm{x^{t+1} - \hat x^{t+1}}^2} = \expec{\norm{x^{t+1} - \hat x^{t} - Q_s(x^{t+1} - \hat x^t)}^2} \leq (1-\delta_s) \expec{\norm{x^{t+1} - \hat x^t}^2}.
    \end{equation}
    Similarly to \cite{choco_sgd} and \cite{EF21-P-and-friends}, we now use the well-known inequality
    \begin{equation}
        \norm{a + b}^2 \leq (1 + \alpha) \norm{a}^2 + (1 + \alpha^{-1}) \norm{b}^2, \quad \forall \alpha > 0.
    \end{equation}
    Applying the definition of $x^{t+1}$ and the previous inequality yields
    \begin{equation}
        \begin{aligned}
            \expec{\norm{x^{t+1} - \hat x^{t+1}}^2} & \leq (1-\delta_s) \expec{\norm{x^{t+1} - \hat x^t}^2}                                                                                                                                        \\
                                                    & \leq (1-\delta_s) \left( 1+\frac{\delta_s}{2} \right) \expec{\norm{x^{t} - \hat x^{t}}^2} + (1-\delta_s) \left( 1 + \frac{2}{\delta_s} \right) \eta_g^2 \expec{\norm{\overline{\Delta}^t}^2} \\
                                                    & \leq \left( 1-\frac{\delta_s}{2} \right) \expec{\norm{x^{t} - \hat x^{t}}^2} + \frac{2}{\delta_s} \eta_g^2 \expec{\norm{\overline{\Delta}^t}^2},
        \end{aligned}
    \end{equation}
    and using induction to substitute the first term concludes the proof.
\end{proof}

\begin{corollary} [Sum of terms from \cref{lemma:hidden-state-biased}] \label{cor:sum-biased-terms}
    From \cref{lemma:hidden-state-biased}, summing for $t = 0, \ldots, T-1$, it follows that
    \begin{equation}
        \sum_{t=0}^{T-1} \expec{ \norm{x^{t} - \hat x^{t}}^2 }  \leq \eta_g^2\frac{2}{\delta_s} \sum_{s=0}^{T-1} \left( 1-\frac{2}{\delta_s} \right)^s \sum_{t=0}^{T-2} \expec{ \norm{\overline{\Delta}^t}^2}.
    \end{equation}
\end{corollary}
\begin{proof}
    \begin{equation}
        \begin{aligned}
            \sum_{t=0}^{T-1} \expec{ \norm{x^{t} - \hat x^{t}}^2 } & \leq  \eta_g^2 \frac{2}{\delta_s} \sum_{t=0}^{T-1} \sum_{s=0}^{t-1} \left(1- \frac{\delta_s}{2 } \right)^{t-1-s} \expec{ \norm{\overline{\Delta}^s}^2 } = \eta_g^2 \frac{2}{\delta_s}  \sum_{s=0}^{T-2} \sum_{t=s+1}^{T-1} \left( 1 - \frac{\delta_s}{2} \right)^{t-1-s} \expec{ \norm{\overline{\Delta}^s}^2 } \\
                                                                   & = \eta_g^2 \frac{2}{\delta_s} \sum_{s=0}^{T-2} \expec{\norm{\overline{\Delta}^s}^2} \sum_{t=s+1}^{T-1} \left( 1 - \frac{\delta_s}{2} \right)^{t-1-s} \leq \eta_g^2 \frac{2}{\delta_s} \sum_{s=0}^{T-2} \expec{\norm{\overline{\Delta}^s}^2} \sum_{t=0}^{T-1} \left( 1 - \frac{\delta_s}{2} \right)^t.
        \end{aligned}
    \end{equation}
\end{proof}

\subsection{Bounding the server step}
\label{sec:bounding-server-step}
The goal of this section is to obtain the following bound on the server step
\begin{equation}
    \expec{\norm{\overline{\Delta}^t }^2} \leq \left(\frac{2-\delta_c}{K}\right)\beta_P \left(\sigma_{\ell}^2 + 2G\right) + \left(1 + \frac{1-\delta_c}{K}\right)P  \sum_{p=0}^{P-1}   (\eta^{(p)}_{\ell})^2 \expec{\norm{\frac{1}{N} \sum_{n = 1}^N \nabla F_n(y^{t-\tau_n}_{n,p}) }^2},
\end{equation}
which is obtained using the same techniques as in the previous sections.
We exploit three key properties: the unbiasedness of the client quantizer, the unbiasedness of the stochastic gradient, and the fact that the expected global gradient is the same as the expected local gradient (choosing clients uniforly at random).
Each property is indicated when used.

\begin{proof}
    First, we exploit the unbiasedness of the client quantizer, and we apply \cref{lemma:sum-unbiasedly-quantized-bound} as follows
    \begin{equation}
        \begin{aligned}
            \expec{ \norm{ \overline{\Delta}^t }^2 }
             & = \expec{\frac{1}{K^2}  \norm{\sum_{k\in \calS_t} Q_c \left( \sum_{p=0}^{P-1} \eta^{(p)}_{\ell} g_k(y^{t-\tau_k}_{k,p}) \right) }^2 }                                                                                                            \\
             & \leq \frac{1}{K^2} \expec{\norm{\sum_{k\in \calS_t} \sum_{p=0}^{P-1} \eta^{(p)}_{\ell} g_k(y^{t-\tau_k}_{k,p})}^2} + \frac{1-\delta_c}{K^2} \expec{ \sum_{k\in \calS_t}  \norm{\sum_{p=0}^{P-1} \eta^{(p)}_{\ell} g_k(y^{t-\tau_k}_{k,p}) }^2 }.
        \end{aligned}
    \end{equation}
    Then, we apply the unbiasedness of the stochastic gradient, and we add and subtract the expectation of our terms, $\pm \nabla F_k(y^{t-\tau_k}_{k,p})$, to obtain
    \begin{equation}
        \begin{aligned}
            \expec{ \norm{ \overline{\Delta}^t }^2 }
             & \leq \frac{1}{K^2} \expec{\norm{\sum_{k\in \calS_t} \sum_{p=0}^{P-1} \eta^{(p)}_{\ell} \nabla F_k(y^{t-\tau_k}_{k,p})}^2} + \frac{1-\delta_c}{K^2} \expec{\sum_{k\in \calS_t} \norm{\sum_{p=0}^{P-1} \eta^{(p)}_{\ell} \nabla F_k(y^{t-\tau_k}_{k,p}) }^2} \\
             & + \frac{1}{K}\beta_P \sigma_{\ell}^2 + \frac{1-\delta_c}{K} \beta_P \sigma_{\ell}^2.
        \end{aligned}
    \end{equation}
    Now, we use the fact that client $k$ is sampled uniformly at random from all clients, and we can ensure
    \begin{equation}
        \expec{ \nabla F_k(y^{t-\tau_k}_{k,p})} = \mathbb{E}_{\mathcal{H}} \left[ \frac{1}{N} \sum_{n = 1}^N \mathbb{E}_{F_n \mid n \sim[N]} \nabla F_n(y^{t-\tau_n}_{n,p}) \right] = \expec{\frac{1}{N} \sum_{n = 1}^N \nabla F_n(y^{t-\tau_n}_{n,p})}.
    \end{equation}
    Thus, again applying the adding and subtracting technique, this time with $\pm \frac{1}{N} \sum_{n = 1}^N \nabla F_n(y^{t-\tau_n}_{n,p}) $,
    \begin{equation}
        \begin{aligned}
            \expec{ \norm{ \overline{\Delta}^t }^2 }
             & \leq \frac{1}{K^2} \expec{\norm{\sum_{k\in \calS_t} \sum_{p=0}^{P-1} \eta^{(p)}_{\ell} \frac{1}{N} \sum_{n = 1}^N \nabla F_n(y^{t-\tau_n}_{n,p}) }^2} + \frac{1-\delta_c}{K^2} \expec{\sum_{k\in \calS_t} \norm{\sum_{p=0}^{P-1} \eta^{(p)}_{\ell} \frac{1}{N} \sum_{n = 1}^N \nabla F_n(y^{t-\tau_n}_{n,p}) }^2} \\
             & + \frac{2-\delta_c}{K^2} \expec{ \sum_{k\in \calS_t} \sum_{p=0}^{P-1} \norm{ \eta^{(p)}_{\ell} \left( \nabla F_k(y^{t-\tau_k}_{k,p}) - \frac{1}{N} \sum_{n = 1}^N  \nabla F_n(y^{t-\tau_n}_{n,p}) \right) }^2}                                                                                                    \\
             & + \frac{2 - \delta_c}{K} \beta_P \sigma_{\ell}^2.
        \end{aligned}
    \end{equation}
    Furthermore, applying the Cauchy-Schwarz and using the bounded gradient from \cref{ass:bounded-and-l-smooth-loss-gradient} to the third term yields
    \begin{equation}
        \begin{aligned}
            \expec{ \norm{ \overline{\Delta}^t }^2 }
             & \leq \frac{1}{K^2} \expec{\norm{\sum_{k\in \calS_t} \sum_{p=0}^{P-1} \eta^{(p)}_{\ell} \frac{1}{N} \sum_{n = 1}^N  F_n(y^{t-\tau_n}_{n,p}) }^2} + \frac{1-\delta_c}{K^2} \expec{\sum_{k\in \calS_t} \norm{\sum_{p=0}^{P-1} \eta^{(p)}_{\ell} \frac{1}{N} \sum_{n = 1}^N  F_n(y^{t-\tau_n}_{n,p}) }^2} \\
             & + \frac{2-\delta_c}{K} \beta_P \left(\sigma_{\ell}^2 + 4G\right)                                                                                                                                                                                                                                      \\
        \end{aligned}
    \end{equation}

    Finally, using Cauchy-Schwarz to bound the norm of the sum gives us our desired bound.
\end{proof}


\end{document}